\documentclass[11pt, letterpaper]{cmu}

\usepackage[all]{hypcap}
\usepackage[comma,numbers,sort,compress]{natbib}
\usepackage{hyperref}[citecolor=magenta]

\hypersetup{
    colorlinks = true,
    citecolor = {magenta},
}

\usepackage{microtype}
\usepackage{graphicx}
\usepackage{subfigure}
\usepackage{booktabs}
\usepackage{float}

\usepackage{amsmath}
\usepackage{amssymb}
\usepackage{mathtools}
\usepackage{amsthm}
\usepackage{mathrsfs}
\usepackage{nicefrac}
\usepackage{dsfont}
\usepackage{enumitem}
\usepackage{float}

\usepackage{xspace}
\usepackage[capitalize,noabbrev]{cleveref}
\usepackage{subcaption}
\usepackage{wrapfig}
\usepackage{listings}

\usepackage{bbm}
\usepackage{fleqn, tabularx}
\usepackage{multirow}
\usepackage[export]{adjustbox}
\usepackage{colortbl}

\usepackage{algpseudocode}
\usepackage{setspace}

\usepackage{color}
\definecolor{deepblue}{rgb}{0,0,0.5}
\definecolor{deepred}{rgb}{0.6,0,0}
\definecolor{deepgreen}{rgb}{0,0.5,0}

\definecolor{rliableolive}{HTML}{BBCC33}
\definecolor{rliableblue}{HTML}{77AADD}
\definecolor{rliablered}{HTML}{EE8866}
\definecolor{myblue}{rgb}{0.1,0.4,0.9}
\definecolor{lightblue}{rgb}{0.22,0.45,0.70}
\definecolor{figurebackground}{HTML}{F9F8F3}

\usepackage{pifont}
\usepackage{soul}
\definecolor{highlightmistake}{RGB}{255, 179, 179}
\definecolor{highlightcorrect}{RGB}{179, 255, 179}

\usepackage[textsize=tiny]{todonotes}
\usepackage{algorithm}
\usepackage{bm}

\usepackage[skins,theorems]{tcolorbox}
\Crefformat{equation}{#2Eq.\;(#1)#3}
\Crefformat{figure}{#2Figure #1#3}
\Crefformat{assumption}{#2Assumption #1#3}
\Crefname{assumption}{Assumption}{Assumptions}

\usepackage{crossreftools}
\tcbset{
  aibox/.style={
    width=\linewidth,
    top=8pt,
    bottom=4pt,
    colback=figurebackground,
    colframe=black,
    colbacktitle=black,
    enhanced,
    center,
    attach boxed title to top left={yshift=-0.1in,xshift=0.15in},
    boxed title style={boxrule=0pt,colframe=white,},
  }
}
\newtcolorbox{AIbox}[2][]{aibox,title=#2,#1}

\newcommand{\ptp}{\texttt{PTP}}
\newcommand{\bpp}{\texttt{BPP}}

\usepackage{amsmath,amsfonts,bm}

\def\eqref#1{equation~\ref{#1}}

\def\1{\bm{1}}

\DeclareMathAlphabet{\mathsfit}{\encodingdefault}{\sfdefault}{m}{sl}
\SetMathAlphabet{\mathsfit}{bold}{\encodingdefault}{\sfdefault}{bx}{n}

\title{\bpp: Long-Context Robot Imitation Learning by Focusing on Key History Frames}

\author[1,2]{Max Sobol Mark}
\author[2]{\!Jacky Liang}
\author[2]{\!Maria Attarian}
\author[2]{\!Chuyuan Fu}
\author[2]{\!Debidatta Dwibedi}
\author[1$\dagger$]{Aviral Kumar\!}
\author[2,3$\dagger$]{Dhruv Shah\!}

\affil[1]{Carnegie Mellon University}
\affil[2]{Google DeepMind}
\affil[3]{Princeton University}

\correspondingauthor{Max Sobol Mark (maxsobolmark@cmu.edu). ~~~~ \emph{Work done during an internship at GDM.} ~~~$^\dagger$\emph{Equal advising.} \\ \textbf{Project website with videos:} \href{https://bigpicturepolicies.github.io/}{https://bigpicturepolicies.github.io/}}

\begin{abstract}
\end{abstract}

\begin{document}

\maketitle

\vspace{-1.2cm}
\begin{center}
\vspace{-0.4cm}
    \begin{tcolorbox}[
        width=0.99\textwidth,
        center,
        colback=figurebackground,
        boxrule=0pt,
        arc=10pt,
        left=6pt,
        right=6pt,
        top=4pt,
        bottom=4pt,
        boxsep=0pt,
        enhanced,
        colframe=figurebackground
    ]
    \centering
    \vspace{-0.1cm}
    \includegraphics[width=\textwidth]{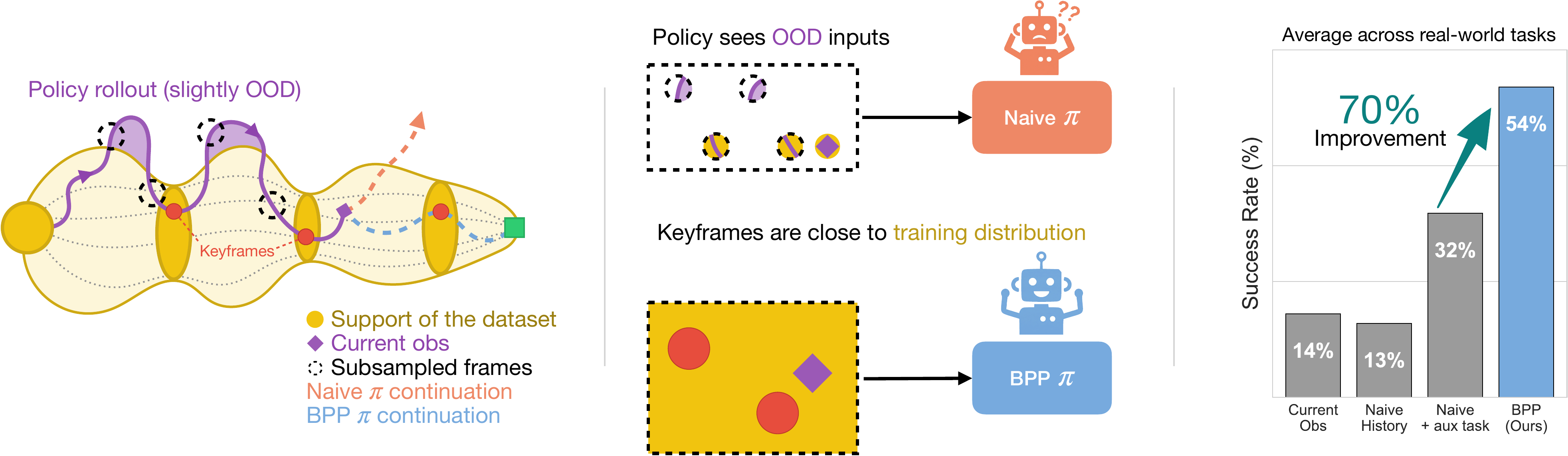}
    \end{tcolorbox}
    \refstepcounter{figure}
    \label{fig:teaser}
    \noindent
    \begin{minipage}{0.99\textwidth}
        \footnotesize \textbf{Figure~1.} \textbf{Big Picture Policies (BPP).} Across a suite of non-Markovian tasks, BPP (left) uses an off-the-shelf vision-language model (VLM) to identify task-relevant keyframes from history, enabling robust performance compared to memoryless and na\"ive history-conditioned baselines (right).
    \end{minipage}
\end{center}

\vspace{-0.2cm}

{\absfont \textbf{Abstract:}     Many robot tasks require attending to the history of past observations. For example, finding an item in a room requires remembering which places have already been searched. However, the best-performing robot policies typically condition only on the current observation, limiting their applicability to such tasks.
    Na\"ively conditioning on past observations often fails due to spurious correlations: policies latch onto incidental features of training histories that do not generalize to out-of-distribution trajectories upon deployment.
    We analyze why policies latch onto these spurious correlations and find that this problem stems from limited coverage over the space of possible histories during training, which grows exponentially with horizon. Existing regularization techniques provide inconsistent benefits across tasks, as they do not fundamentally address this coverage problem.
    Motivated by these findings, we propose \textbf{Big Picture Policies} (BPP), an approach that conditions on a minimal set of meaningful keyframes detected by a vision-language model. By projecting diverse rollouts onto a compact set of task-relevant events, BPP substantially reduces distribution shift between training and deployment, without sacrificing expressivity. We evaluate BPP on four challenging real-world manipulation tasks and three simulation tasks, all requiring history conditioning. BPP achieves 70\% higher success rates than the best comparison on real-world evaluations.
}

\vspace{-0.2cm}
\section{Introduction}
\vspace{-0.2cm}

Imitation learning has become a central paradigm for training robotic policies. However, the most successful approaches today typically lack \emph{memory}, conditioning only on the current observation~\citep{kim2025openvla,team2025gemini2,intelligence2025pi_,kim2026cosmos,zitkovich2023rt} or a short temporal window~\citep{ghosh2024octo,brohan2022rt}. While effective on many benchmarks, such policies struggle in settings where the robot must remember what it has already done, seen, or what remains to be completed. Consider tasks such as searching for an object in a cluttered scene, adding a precise number of ingredients, or executing multi-stage procedures where early mistakes are not easily observed later. Here, the current observation is insufficient to determine the correct action; success instead requires reasoning over past events. Motivated by this, we study training policies to effectively leverage history.

A natural approach for learning memory-based policies is to condition policies directly on sequences of past observations. Yet, despite its appeal, prior work has consistently found that na\"ively adding history often \emph{hurts} performance~\citep{wen2020fighting,wen2021keyframe,diffusionpolicy_chi2023,zhengtracevla,torne2025learning}, a phenomenon we also observe very prominently in the real world. Rather than improving robustness, history-conditioned policies frequently fail even on simple tasks. These failures arise because policies latch onto incidental features (i.e., \emph{spurious correlations}) of training histories that explain expert actions but do not generalize. Once the policy deviates from the expert distribution, it encounters unseen histories and errors compound.

In theory, this problem stems from a fundamental challenge of \emph{coverage}: the space of possible observation histories grows exponentially with horizon, which makes it impossible to cover during training data collection. Coverage is especially limited under current robotic data collection paradigms, which rely primarily on human tele-operators gathering ``near-expert'' demonstrations. As a result, although history-conditioned policies can predict the underlying \emph{oracle} task state accurately on the training set, they fail catastrophically on histories encountered during deployment. Existing approaches attempt to mitigate the spurious correlations through regularization~\citep{seo2023regularized}, auxiliary objectives~\citep{torne2025learning}, or architectural constraints~\citep{zhengtracevla}. While these techniques can help in some cases, our experiments show their benefits are inconsistent and often task-dependent as they do not fundamentally address coverage. In essence, they modify the loss function without addressing the underlying lack of coverage over histories.

These findings suggest that effective history conditioning requires directly addressing the challenge of coverage. {But \emph{how can we improve coverage without collecting more data?}} Our core idea is to train policies with a different input space than raw observation histories. In particular, we replace full action histories with a small number of past moments that are most relevant for predicting actions. In many non-Markovian tasks, successful behavior depends on identifying a few behaviorally salient events, or \emph{keyframes}, such as whether an object has been grasped, a subgoal has been completed, or a failure has occurred. Representing history through these semantic keyframes increases overlap between training and test-time policy inputs by reducing the size of the effective input history space, while preserving all the task-relevant information needed for accurate action prediction. This enables robust generalization despite limited coverage over all histories in the data.

In this paper, we instantiate this idea via a concrete approach we call \textbf{Big Picture Policies (\bpp)}, which trains robotic policies conditioned on semantic keyframes. \bpp{} leverages an off-the-shelf vision-language model (VLM) to identify task-relevant keyframes via simple question answering, projecting long and diverse trajectory histories into a compact set of informative inputs for policy learning. We evaluate \bpp{} on four challenging real-world bimanual manipulation tasks and three simulated tasks, all of which require memory to succeed. Across all settings, \bpp{} substantially outperforms strong history-conditioned and memoryless baselines, achieving up to 70\% higher success rates in real-world evaluations. Our contributions are twofold: \textbf{(i)} we show that the central challenge in history-conditioned policy learning is coverage over histories, and that regularization-based approaches provide inconsistent, task-dependent improvements without resolving the core underlying issue; and \textbf{(ii)} we introduce \bpp{},
a simple yet effective method that leverages VLMs to identify task-relevant keyframes, achieving state-of-the-art performance with improved data efficiency and reduced training time.

\vspace{-0.2cm}
\section{Related Work}
\label{sec:related_work}
\vspace{-0.2cm}

\textbf{History in robot learning policies.}
Most high-performing robot policies predict actions conditioned only on the current observation~\citep{kim2025openvla,intelligence2025pi_,zitkovich2023rt,team2025gemini,team2025gemini2,kim2026cosmos}. This raises a natural question: is history conditioning actually necessary? Much of robot learning work evaluates on tasks that do not require memory, though some prior work studies settings where history is required~\citep{torne2025learning,sridhar2025memer} or beneficial~\citep{zhang2018deep,brohan2022rt,zhengtracevla}. In contrast, on-policy deep RL routinely uses memory, via frame stacking or recurrent networks, to handle partial observability~\citep{mnih2016asynchronous,andrychowicz2020learning}.  
History is less common in robot imitation learning because adding observation histories often hurts performance~\citep{wen2020fighting,wen2021keyframe,diffusionpolicy_chi2023,zhengtracevla,torne2025learning}, increasing spurious correlations and causal confusion under rollout distribution shift~\citep{de2019causal,bansal2018chauffeurnet,swamy2022sequence}. Our goal in this paper is to address these spurious correlations.

\textbf{Addressing spurious correlations.}
A well-studied failure mode in imitation from observation histories is the \emph{copycat} problem~\citep{wen2020fighting}, also known as causal confusion~\citep{de2019causal}, where policies overfit to past actions implicitly encoded in observation sequences rather than responding to the current state. Prior work addressed this issue by upweighting loss on keyframes where actions change significantly~\citep{wen2021keyframe}, discovering keyframe structure for planning~\citep{pertsch2020keyframing}, or applying information bottlenecks to suppress past-action leakage~\citep{seo2023regularized}. However, copycat is not the only source of spurious correlations in long-context policies: even without explicit action histories, extended observation sequences create opportunities for policies to latch onto irrelevant temporal patterns.

Beyond loss reweighting, prior work has proposed auxiliary tasks to address spurious correlations. \citet{torne2025learning} introduce past-token prediction as a self-supervised objective and employ a test-time selection mechanism to enforce temporal consistency, which we find is effective only under certain task conditions. Another line of work avoids conditioning the low-level policy on raw observation histories altogether. TraceVLA~\citep{zhengtracevla} augments observations with a visual trace of the robot’s past trajectory, while MemER~\citep{sridhar2025memer} retrieves relevant past experiences via a language-based memory and delegates action prediction to a memoryless policy.  
We argue that such approaches are limited: TraceVLA’s visual trace cannot capture object-level or scene history, and MemER’s language-only memory limits expressivity and assumes a clean sub-task decomposition, \textit{e.g.,} if memory is required within a single sub-task, the low-level policy cannot complete it. In contrast, our method conditions the low-level policy directly on semantically meaningful frames from history, preserving expressivity while mitigating spurious correlations.

\textbf{Pre-trained models in robotics.}
Vision-language models (VLMs) pre-trained on internet-scale data have been used for robotics in several ways. Prior work fine-tunes VLMs directly for action prediction, yielding vision-language-action (VLA) models~\citep{kim2025openvla,zitkovich2023rt,intelligence2025pi_,ghosh2024octo}. Other approaches keep the VLM frozen and use it as a high-level planner issuing language commands to a low-level policy~\citep{ahn2022can,liang2023code,wang2024vlm}, or as a reward model, success detector, or value function for reinforcement learning~\citep{du2023vision,ma2024vision,baumli2023vision,yang2024robot}. 
Our work falls into a distinct category: using a VLM to identify semantically meaningful moments in the observation history, guiding what the policy should attend to rather than what actions it should take. While some prior work (e.g., \citep{black2023zero}) uses VLMs to propose ``important'' states, these are typically treated as future sub-goals. To our knowledge, prior work has not used VLMs to filter observation histories to improve learning signals.

\vspace{-0.2cm}
\section{What Makes History-Conditioned Imitation Learning Particularly Challenging?}
\label{sec:analysis}
\vspace{-0.2cm}

We first examine why history-conditioned imitation is difficult, and whether practices now standard in robot imitation learning, such as expressive policy classes (e.g., diffusion~\citep{diffusionpolicy_chi2023} or flow policies~\citep{flow_lipman2023}) with action chunking~\citep{zhao2023learning} can mitigate these issues. Conventional wisdom attributes challenges of history conditioning to \emph{spurious correlations}: when conditioned on histories, policies may latch onto features that explain training actions but fail to generalize to test-time histories. To assess this, we first study standard (na\"ive) history conditioning with an action-chunking diffusion policy, focusing on spurious correlations.

\textbf{Notation.} We consider partially observable decision-making problems with latent states $s_t \in \mathcal{S}$, observations $o_t \in \mathcal{O}$, and actions in $\mathcal{A}$. Because tasks are often non-Markovian with respect to $o_t$, we define the observation history as $h_t = (o_{t-k}, \dots, o_t)$. Given a dataset of $N$ expert demonstrations $\mathcal{D} = \{\tau_i\}_{i=1}^{N}$, our objective is to learn a history-conditioned policy $\pi_\theta: \mathcal{O}^{k+1} \to \Delta(\mathcal{A})$ via imitation learning.

\vspace{-0.1cm}
\subsection{History with Action-Chunking Diffusion Policies}
\vspace{-0.1cm}

\begin{figure*}[t]
    \centering
    \vspace{-0.4cm}
    \begin{tcolorbox}[
        colback=figurebackground,
        boxrule=0pt,
        arc=10pt,
        left=6pt,
        right=6pt,
        top=4pt,
        bottom=4pt,
        boxsep=0pt,
        enhanced,
        colframe=figurebackground
    ]
    \begin{minipage}[c]{0.12\textwidth}
        \centering
        {\small\textsf{\textbf{\color[HTML]{8C1515}Aloha 2}}}\\
        {\small\textsf{\textbf{\color[HTML]{8C1515}(real)}}}
    \end{minipage}%
    \hfill
    \begin{minipage}[c]{0.85\textwidth}
        \centering
        \begin{minipage}[b]{0.235\textwidth}
            \centering
            \includegraphics[width=\textwidth]{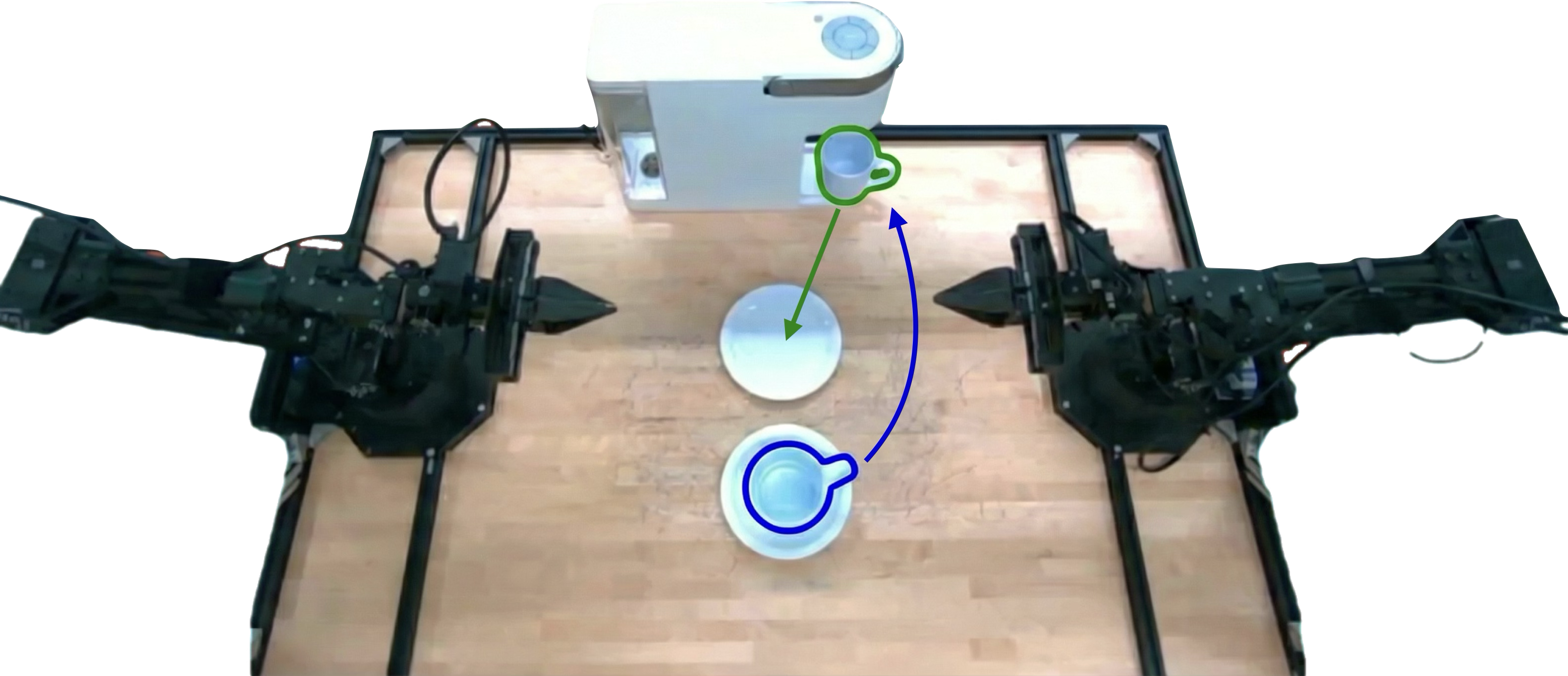}
            {\scriptsize A. Mug Replacement}
        \end{minipage}\hfill
        \begin{minipage}[b]{0.235\textwidth}
            \centering
            \includegraphics[width=\textwidth]{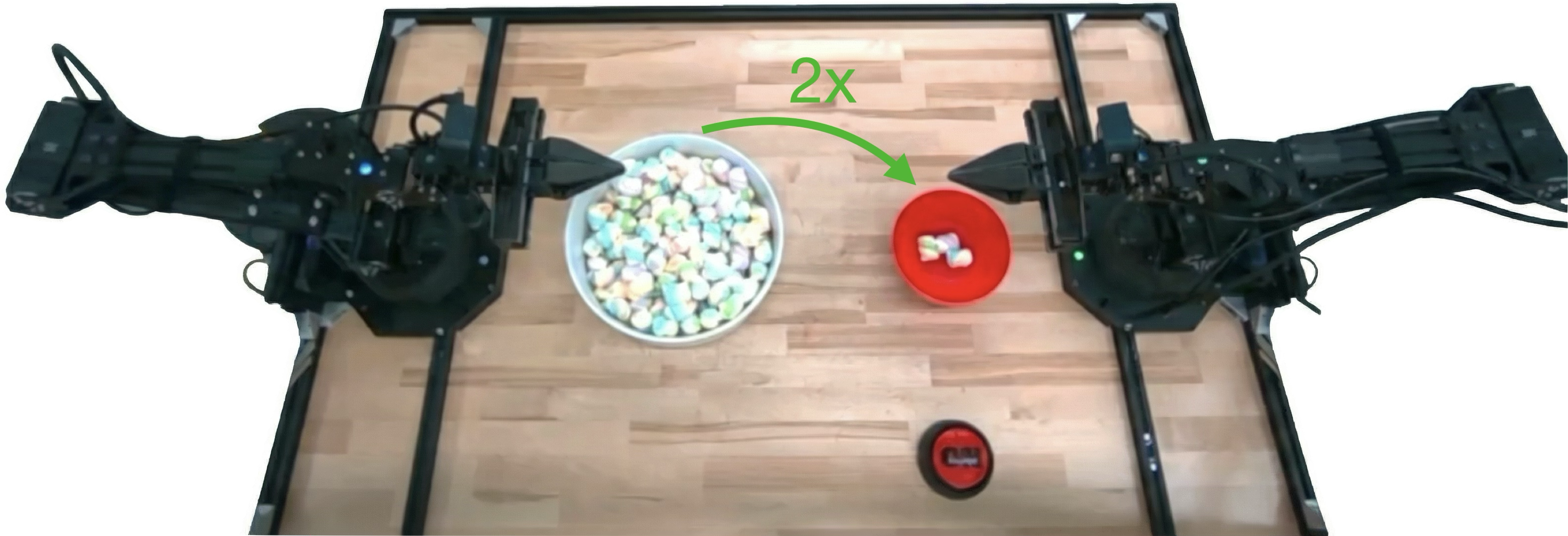}
            {\scriptsize B. Marshmallows}
        \end{minipage}\hfill
        \begin{minipage}[b]{0.235\textwidth}
            \centering
            \includegraphics[width=\textwidth]{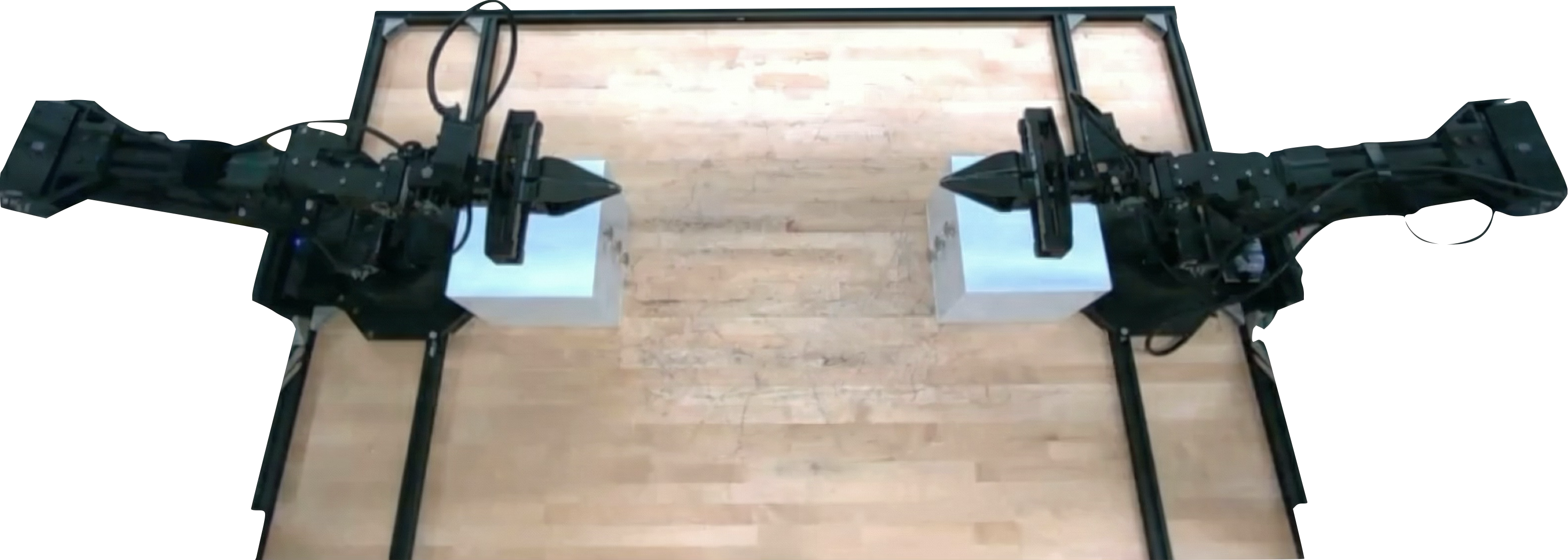}
            {\scriptsize C. Drawer Search}
        \end{minipage}\hfill
        \begin{minipage}[b]{0.235\textwidth}
            \centering
            \includegraphics[width=\textwidth]{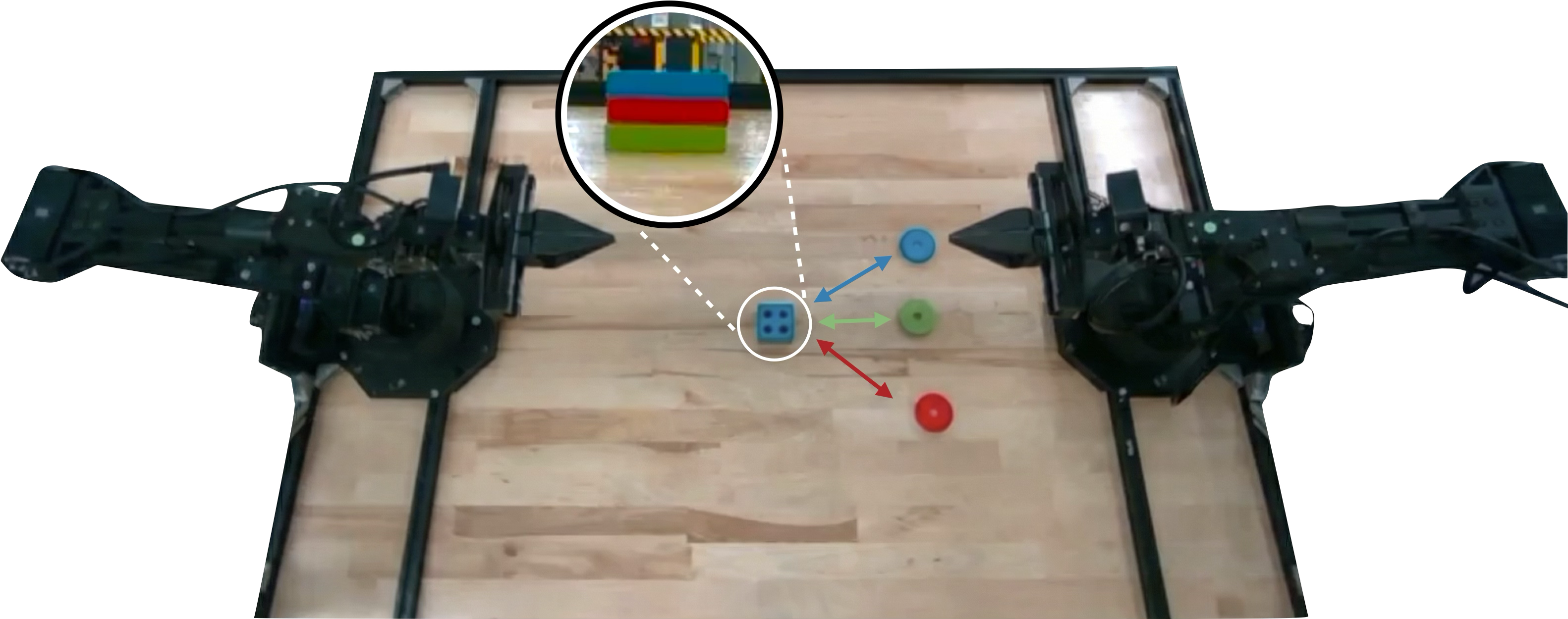}
            {\scriptsize D. Stacking Puzzle}
        \end{minipage}
    \end{minipage}
    
    \vspace{2pt}
    {\color{gray}\noindent\rule{\textwidth}{0.4pt}}\par
    \nointerlineskip
    \vspace{4pt}
    
    \begin{minipage}[c]{0.12\textwidth}
        \centering
        {\small\textsf{\textbf{\color{blue}MuJoCo}}}\\
        {\small\textsf{\textbf{\color{blue}(sim)}}}
        \vspace{6pt}
    \end{minipage}%
    \hfill
    \begin{minipage}[c]{0.85\textwidth}
        \centering
        \begin{minipage}[b]{0.235\textwidth}
            \centering
            \includegraphics[width=\textwidth]{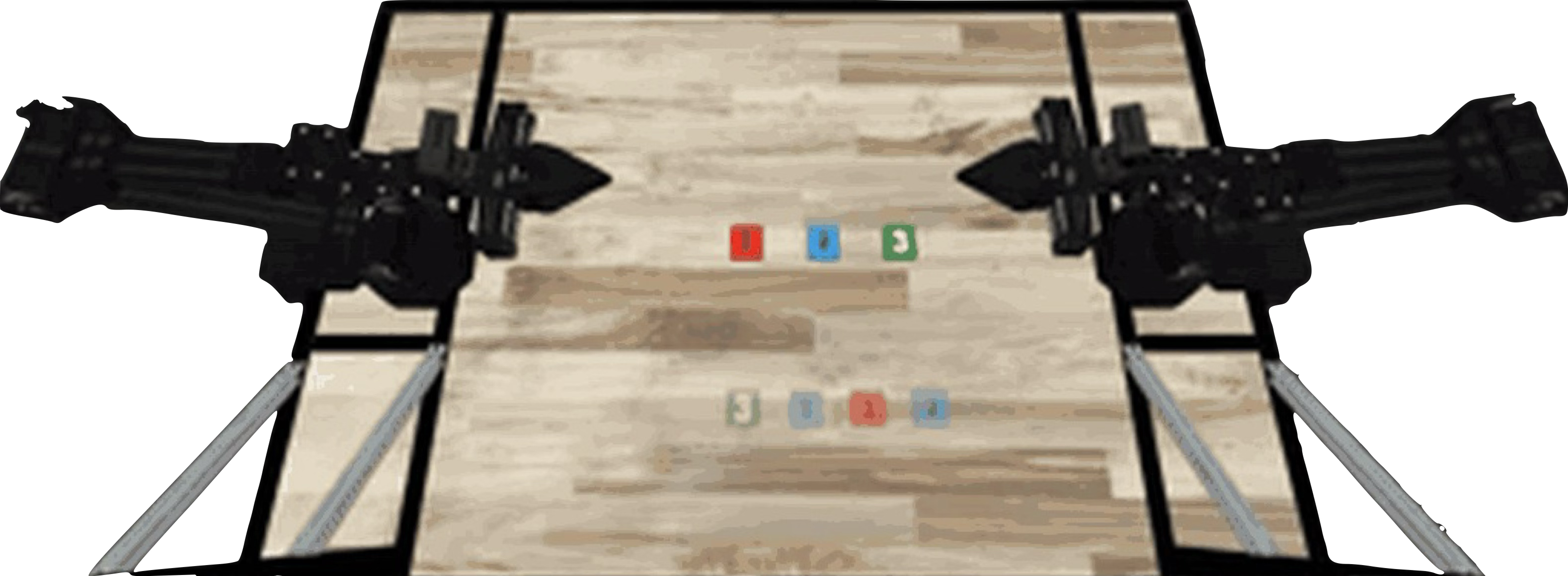}
            {\scriptsize E. Variable Password}
        \end{minipage}\hspace{0.04\textwidth}
        \begin{minipage}[b]{0.235\textwidth}
            \centering
            \includegraphics[width=\textwidth]{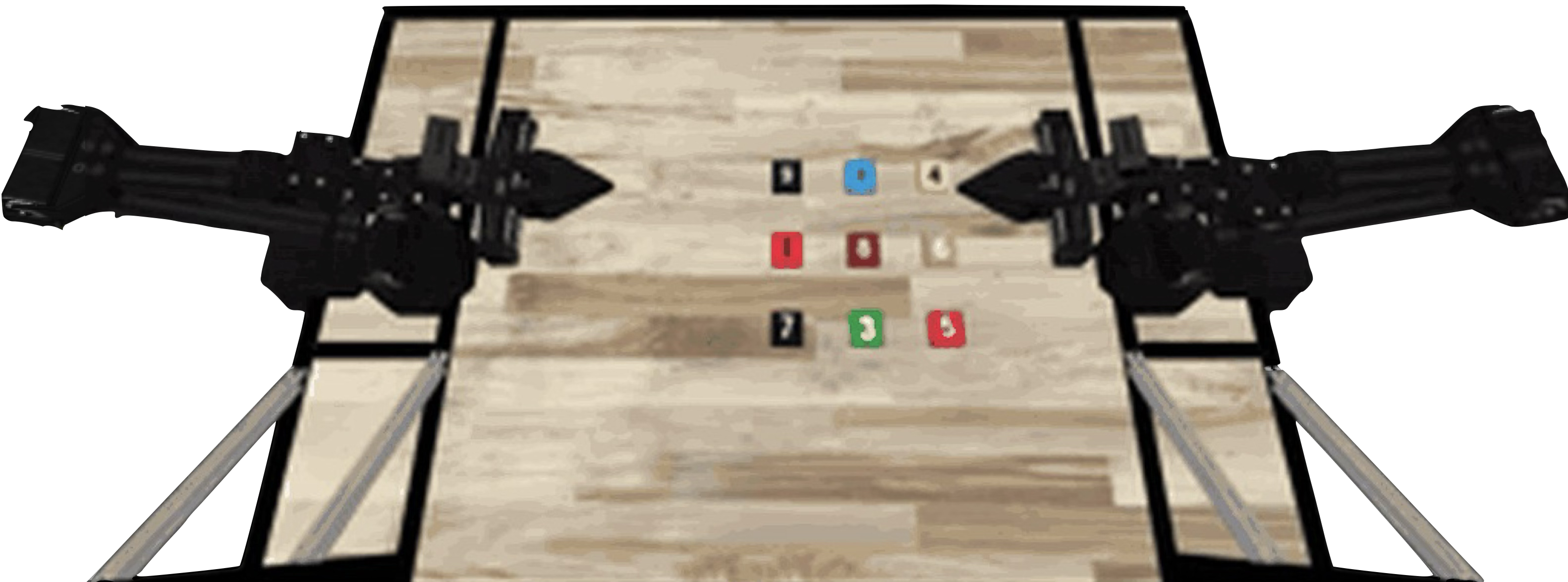}
            {\scriptsize F. Fixed Password}
        \end{minipage}\hspace{0.04\textwidth}
        \begin{minipage}[b]{0.235\textwidth}
            \centering
            \includegraphics[width=\textwidth]{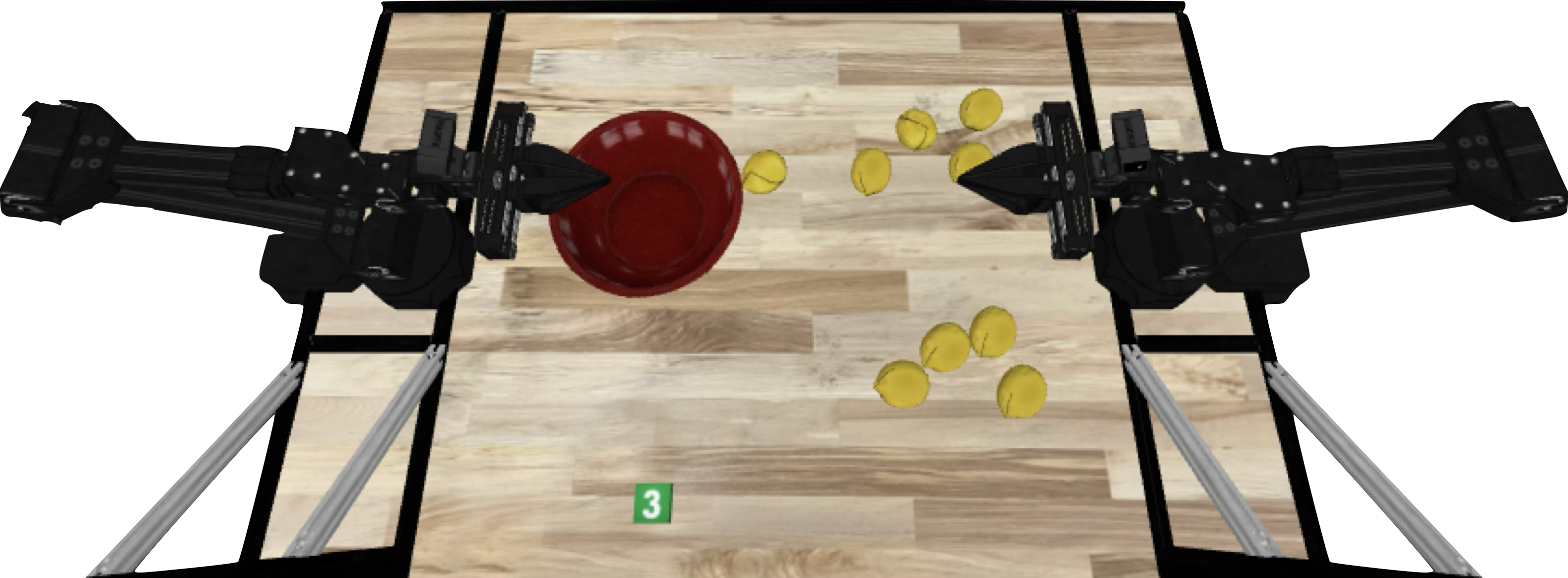}
            {\scriptsize G. Ingredient Insertion}
        \end{minipage}
    \end{minipage}
    \end{tcolorbox}
    \vspace{-0.2cm}
    \caption{\textbf{Benchmark tasks.} We evaluate on 4 real-world tasks (A--D) and 3 simulation tasks (E--G). All tasks require history conditioning for success.}
    \label{fig:environments}
    \vspace{-15pt}
\end{figure*}

We first benchmark action-chunking diffusion policies on a suite of simulation tasks explicitly designed to be unsolvable without leveraging history~(Figure~\ref{fig:environments}):
\begin{itemize}[leftmargin=*,itemsep=4pt,topsep=0pt]
    \item $\mathrm{Ingredient\text{-}Insertion}$: Place two lemons inside a red bowl, then press a button. Once a lemon is in the bowl, it becomes invisible. This mirrors tasks like adding spoonfuls of sugar to coffee, where previous actions must be remembered as they are no longer observable in the current state.
    \item $\mathrm{Fixed\text{-}Password}$: Enter a 7-digit password (1112134) on a $3 \times 3$ grid of buttons. While the password remains the same, button positions are randomized each episode. This task needs ``dynamic memory'': the number of relevant history frames varies depending on the digit being entered.
    \item $\mathrm{Variable\text{-}Password}$: Given a 4-digit password (as a one-hot vector), enter it using three available buttons. Each password contains at least one repeated digit, requiring the policy to track its progress.
\end{itemize}
To establish a lower bound on performance, we run imitation using only the current observation. We then evaluate na\"ive history conditioning, where policy inputs contain a fixed-length, uniformly-strided window of past observations. We chose the history length and stride so that they provide sufficient representation for the optimal policy (see Appendix~\ref{sec:appendix_task_details}). Finally, we establish an upper bound using an \emph{oracle} policy that has access to privileged state information, making the task solvable without history.

Our results (Figure~\ref{fig:sim_combined_bar}) reveal a dichotomy: while there is a substantial performance gap between na\"ive history conditioning and the oracle, the na\"ive method does not suffer from the poor performance as reported by prior work~\citep{de2019causal, wen2020fighting, torne2025learning}. In the following section, we investigate both what prevents catastrophic failure for na\"ive history conditioning and the fundamental factors that limit its performance.

\begin{figure*}[t]
    \centering
    \vspace{-0.4cm}
    \begin{tcolorbox}[
        colback=figurebackground,
        boxrule=0pt,
        arc=10pt,
        left=6pt,
        right=6pt,
        top=4pt,
        bottom=4pt,
        boxsep=0pt,
        enhanced,
        colframe=figurebackground
    ]
    \centering
    \includegraphics[width=\textwidth]{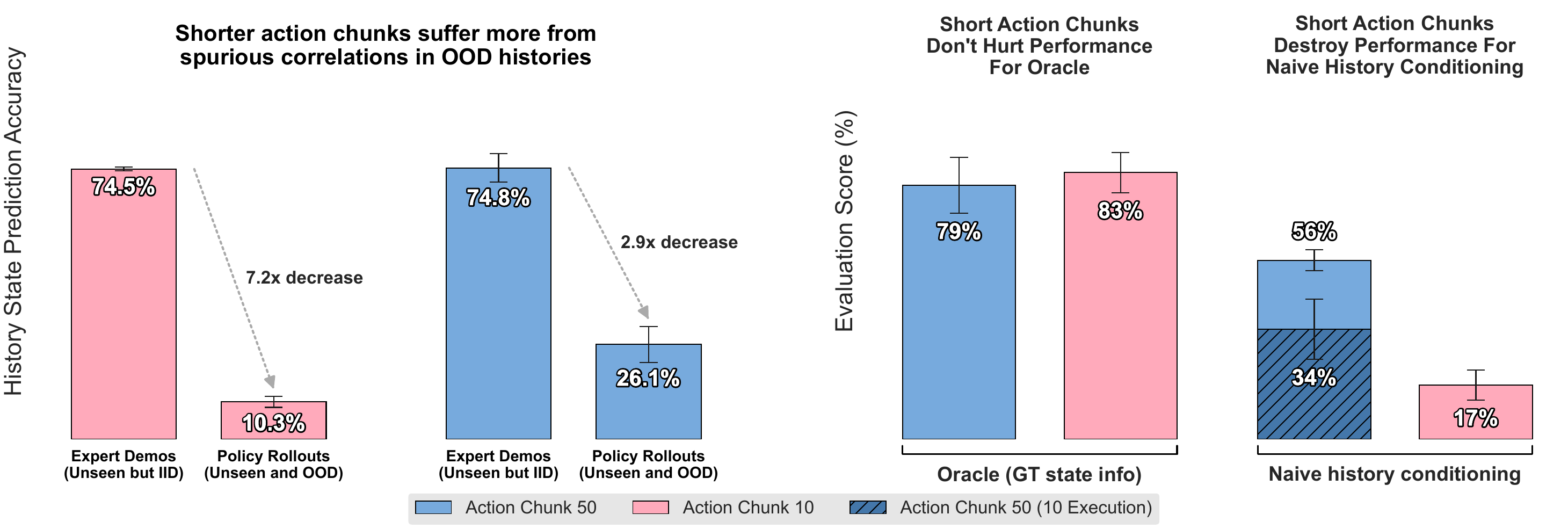}
    \end{tcolorbox}
    \vspace{-0.3cm}
    \caption{\textbf{Predicting longer action chunks significantly decreases spurious correlations in na\"ive history-conditioned policies.} \textbf{Left:} We compare how well the features learned by the policy predict the history state (in $\mathrm{Fixed\text{-}Password}$, this is the number of buttons pressed so far). Both policies trained with shorter (10) and longer (50) action chunks learn features that predict history state well on in-distribution expert trajectories. However, policies trained with shorter action chunks exhibit a much larger generalization gap to out-of-distribution policy rollouts (7.2$\times$ vs. 2.9$\times$ decrease), indicating greater reliance on spurious correlations. \textbf{Right:} These representational differences translate into significant performance gaps. Shorter action chunks drastically hurt na\"ive history-conditioned policy performance, while having little effect on oracle policies. Notably, training with action chunk 50 but executing with chunk 10 still significantly outperforms training with chunk 10, confirming that longer chunk prediction improves feature learning for history conditioning.}
    \label{fig:action_chunking_combined}
    \vspace{-10pt}
\end{figure*}

\vspace{-0.1cm}
\subsection{Why is Na\"ive History Conditioning Able to Avoid Catastrophic Failure?}
\label{sec:avoiding_catastrophe}
\vspace{-0.1cm}

We now ablate various design components of our policy to understand why na\"ive history conditioning avoids catastrophic failure, despite conventional wisdom suggesting it should fail. We hypothesize that two components play a key role: \textbf{1)} the use of action chunking, and \textbf{2)} the policy architecture. We explain our hypotheses below and perform controlled experiments to understand the role of each component.

\emph{\textbf{(a) Action chunking.}} We hypothesize that predicting temporally extended action chunks suppresses reliance on spurious correlations in observation histories, even though these segments correspond only to future timesteps and not past ones as in~\citet{torne2025learning}. To test this, we evaluate how well policy features encode history state (e.g., the number of buttons pressed in $\mathrm{Fixed\text{-}Password}$) by training a classifier on top of the learned representations. In Figure~\ref{fig:action_chunking_combined} (left), policies with short action chunks accurately predict history state on i.i.d. validation trajectories, but they fail under rollout-induced distribution shift, achieving near-random accuracy. By contrast, longer action chunks yield more robust features that generalize better to policy rollouts (2.9$\times$ vs 7.2$\times$ error increase). However, even with longer chunks, a substantial gap remains between expert and rollout performance, indicating that action chunking mitigates but does not eliminate the issue.

These representational differences directly translate into performance. Figure~\ref{fig:action_chunking_combined} (right) shows that while shortening the predicted chunk length has little effect on the oracle policy, it causes a substantial performance decline in the na\"ive history-conditioned policy. This indicates that action chunking plays a crucial role in mitigating spurious correlations: by enforcing longer-horizon action prediction, it forces the policy to learn features that are more robust to history shift.

\emph{\textbf{(b) Policy architecture.}} A common strategy for reducing the cost of long histories is a multi-stage recipe: the visual encoder is trained with short contexts and then frozen~\citep{torne2025learning}. In Appendix~\ref{sec:appendix_frozen_encoder}, we find that this substantially degrades history conditioning, suggesting that encoder inflexibility contributes to the poor performance observed by prior work~\citep{torne2025learning}. By contrast, our approach trains a single encoder jointly across all timesteps, allowing representations to adapt to extended histories and yielding better performance.

\vspace{-0.1cm}
\subsection{Why Does the Performance Gap to the Oracle Persist?}
\label{sec:gap_persists}
\vspace{-0.1cm}

While appropriate architectural choices prevent na\"ive history conditioning from failing catastrophically, a large performance gap remains to the oracle (Figure~\ref{fig:sim_combined_bar}). We found above that even with long action chunking, the policy's understanding of the underlying state degrades significantly on its own rollouts. This raises a critical question: can we solve this problem by simply adding a better auxiliary regularizer?

\begin{figure}[t]
\centering

\begin{minipage}[t]{0.47\textwidth}
\centering
\vspace{-0.4cm}
\begin{tcolorbox}[
    colback=figurebackground,
    boxrule=0pt,
    arc=10pt,
    left=6pt,
    right=6pt,
    top=4pt,
    bottom=4pt,
    boxsep=0pt,
    enhanced,
    colframe=figurebackground
]
\centering
\includegraphics[width=\linewidth]{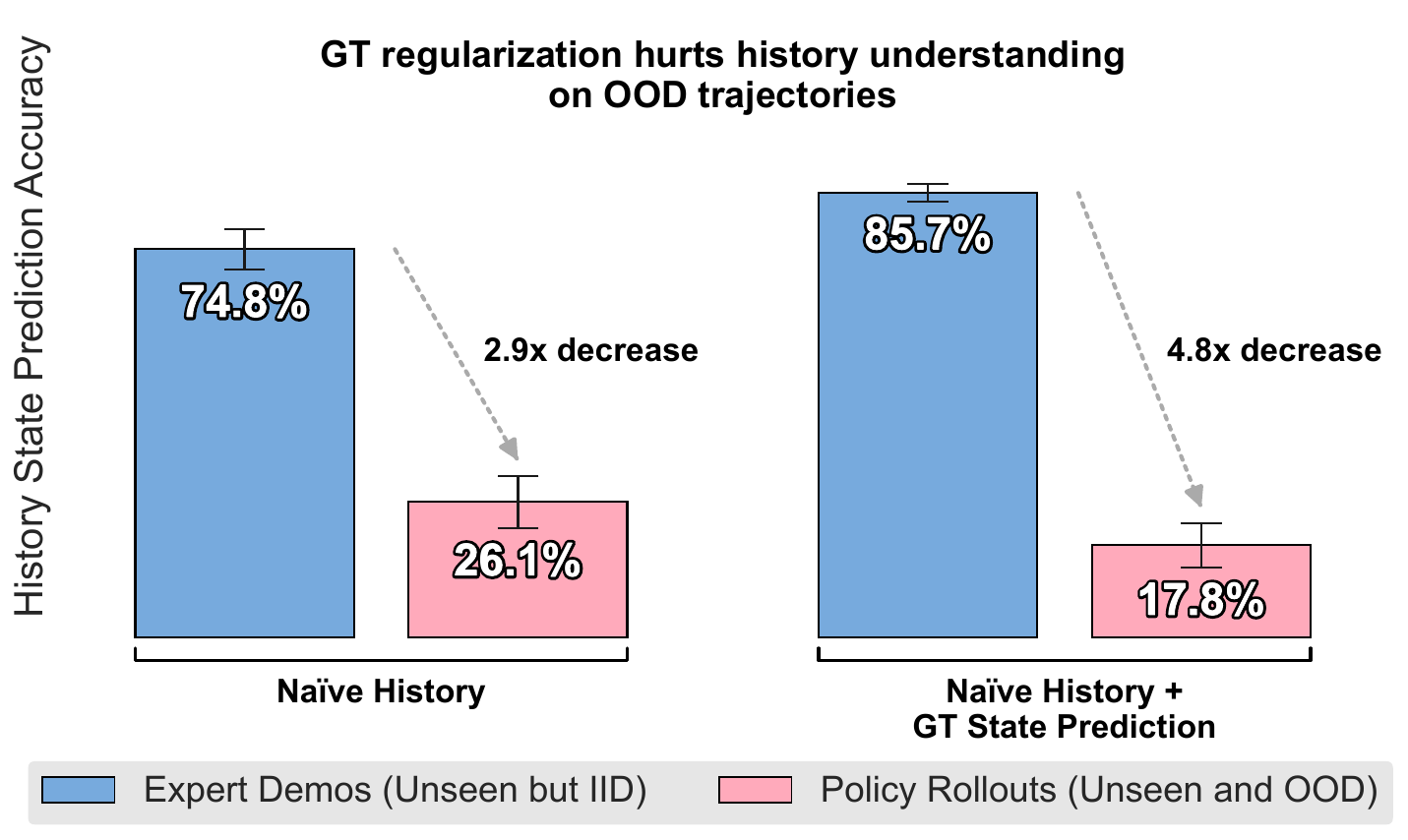}
\end{tcolorbox}
\vspace{-0.4cm}
\caption{%
\textbf{History state prediction regularization hurts history understanding.}
We evaluate whether regularizing the history encoder to predict ground-truth state information (number of buttons pressed so far in $\mathrm{Fixed\text{-}Password}$) improves generalization. While this auxiliary task improves accuracy on unseen expert trajectories, it leads to worse performance on out-of-distribution rollouts, indicating it increases reliance on spurious correlations. This regularization also degrades success rate: $55.5\% \pm 3.3\%$ to $19.0\% \pm 3.3\%$.}
\label{fig:gt_regularization}
\end{minipage}
~\textcolor{gray}{\vline}~
\begin{minipage}[t]{0.47\textwidth}
\centering
\vspace{-0.4cm}
\begin{tcolorbox}[
    colback=figurebackground,
    boxrule=0pt,
    arc=10pt,
    left=6pt,
    right=6pt,
    top=4pt,
    bottom=4pt,
    boxsep=0pt,
    enhanced,
    colframe=figurebackground
]
\centering
\vspace{0.05cm}
\includegraphics[width=\linewidth]{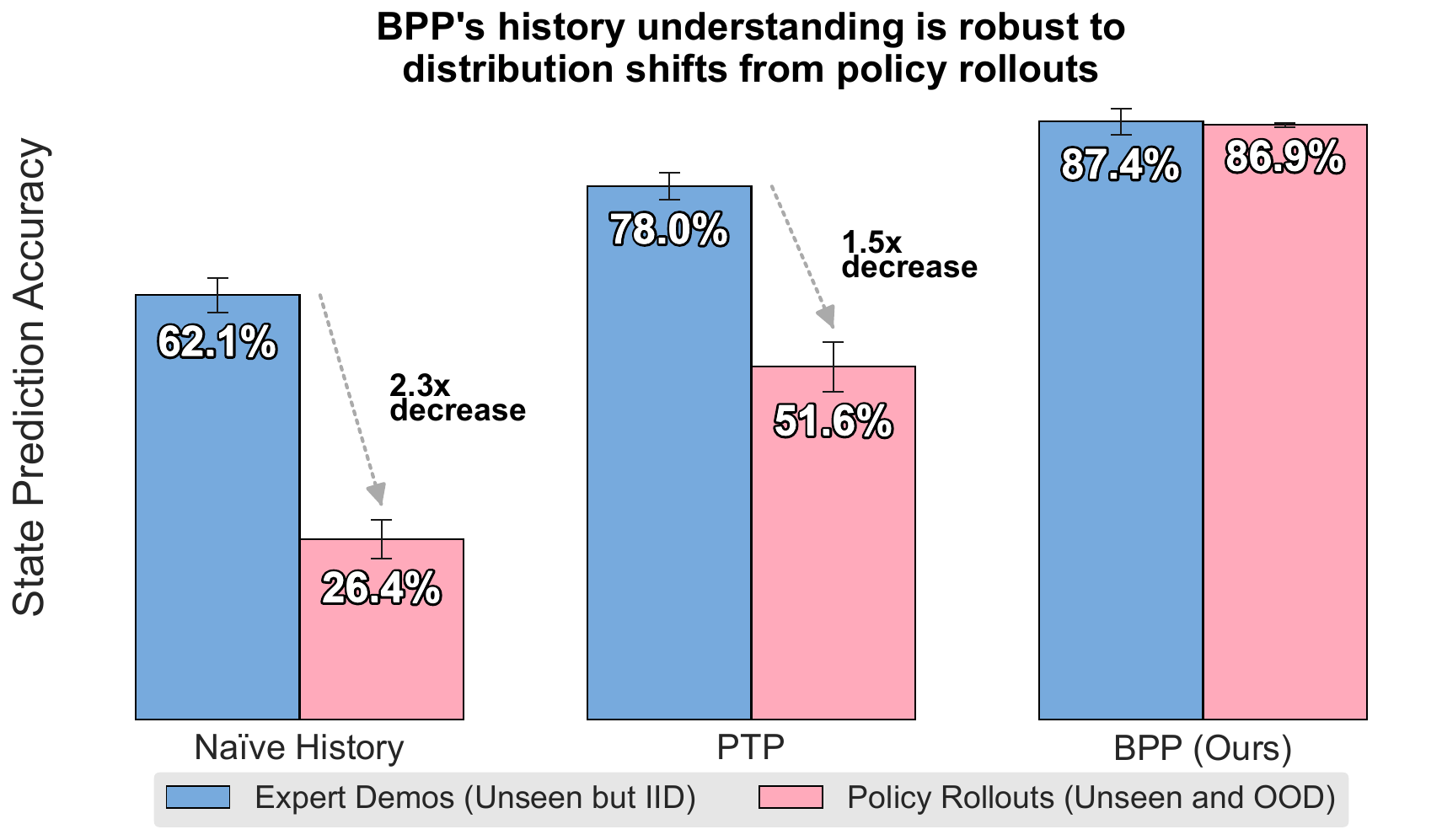}
\end{tcolorbox}
\vspace{-0.35cm}
\caption{%
\textbf{\bpp{} is more robust to the distribution shift between histories appearing in the training data and policy rollouts.}
We evaluate accuracy of state prediction given the observation history, averaged across the fixed-password-entering and ingredient-insertion tasks. Both Na\"ive History Conditioning and \ptp{} suffer big degradations in performance when testing on policy rollouts. Our approach, \bpp{}, which we discuss next, does not suffer from this drop.}
\label{fig:state_prediction_averaged}
\end{minipage}

\vspace{-0.4cm}
\end{figure}

To test this, we construct a ``golden'' auxiliary task: explicitly regularizing the visual encoder to predict 
the ground-truth history state (e.g., the number of buttons pressed so far in $\mathrm{Fixed\text{-}Password}$). Since this loss directly forces the encoder to extract the exact information needed to solve the task, it provides a strong stress-test for auxiliary regularizers. Observe in Figure~\ref{fig:gt_regularization}, adding this auxiliary loss increases accuracy on in-distribution validation data. However, surprisingly, it leads to a decrease in both the understanding of the underlying history state on out-of-distribution rollouts (from 86\% to 18\%) and the policy success rate on deployment (from $56\%$ to $19\%$).

This \emph{negative} result highlights a critical limitation of history conditioning on expert demonstrations: even when the learned features can make correct state predictions on the training distribution, these features become spurious under distribution shift if the training data lacks sufficient coverage of corrective behaviors. This points to \emph{coverage}, rather than feature learning architecture, as the fundamental bottleneck. In the next section, we introduce \textbf{Big Picture Policies (BPP)}, our method that bypasses this limitation. Observe in Figure~\ref{fig:state_prediction_averaged}, while na\"ive and PTP suffer significant degradation in history understanding when evaluated on policy rollouts, BPP does not.

\begin{AIbox}{Takeaways: Lack of coverage of observation histories is a key bottleneck}
\begin{itemize}[leftmargin=*,itemsep=4pt,topsep=0pt]
    \item Larger action chunks and jointly training image encoders on all timesteps help avoid catastrophic failures with na\"ive history conditioning, but history-conditioned policies still lack history \emph{understanding} on the distribution of histories induced by on-policy rollouts.
    \item Regularizing via ``golden'' ground-truth representation can actually \emph{hurt} out-of-distribution performance, indicating that \textbf{coverage}, not architecture or objective, is the bottleneck.
\end{itemize}
\end{AIbox}

\vspace{-0.2cm}
\section{A Simple Recipe for History-Conditioning in Imitation Learning}
\label{sec:methods}
\vspace{-0.2cm}

\begin{figure*}[t]
    \centering
    \vspace{-0.cm}
    \begin{tcolorbox}[
        width=0.95\textwidth,
        colback=figurebackground,
        boxrule=0pt,
        arc=10pt,
        left=6pt,
        right=6pt,
        top=4pt,
        bottom=4pt,
        boxsep=0pt,
        enhanced,
        colframe=figurebackground
    ]
    \centering
    \includegraphics[width=\textwidth]{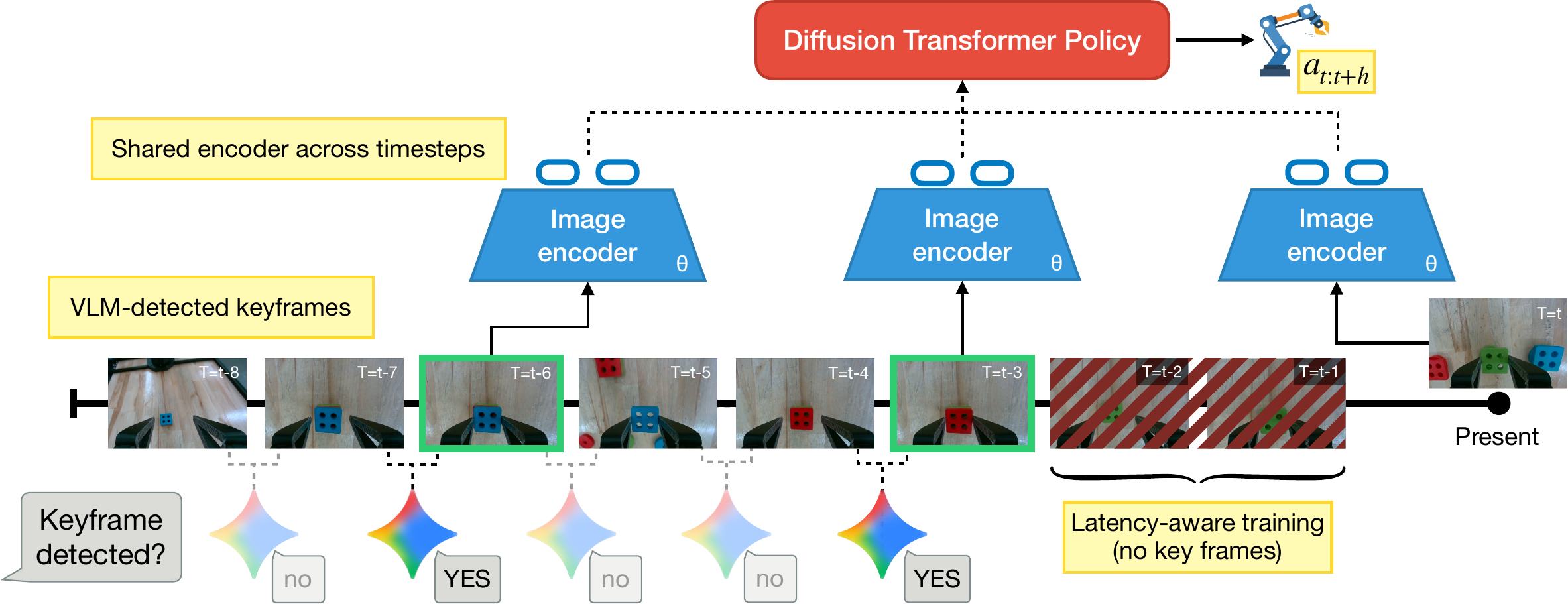}
    \end{tcolorbox}
    \vspace{-0.4cm}
    \caption{\textbf{\bpp{} system architecture.} We condition a standard diffusion transformer policy architecture on a small set of history keyframes. These keyframes are defined by simple, task-specific criteria and are detected using a VLM. We also mask recent histories to account for detection latency.}
    \label{fig:keyframes}
    \vspace{-0.3cm}
\end{figure*}

In the previous section, we showed that na\"ive history conditioning fails due to insufficient coverage over observation histories, so policies learn to rely on spurious correlations. Existing approaches based on regularization do not resolve this issue because they do not address the underlying mismatch between the histories observed during training and those encountered at deployment. This motivates a simple question: \emph{can we improve history conditioning without changing data collection protocols or training objectives?} In this section, we introduce \emph{Big Picture Policies} (\textbf{\bpp}), which address the coverage challenge by changing the \emph{representation} of history rather than the learning algorithm itself (i.e., standard imitation learning without auxiliary regularization). Instead of conditioning policies on full observation sequences, \bpp{} conditions on a \emph{minimal} set of task-relevant \emph{keyframes}. By compressing long and diverse histories into a small number of keyframes, \bpp{} increases the overlap between policy inputs at training and test time, reducing the exposure to spurious correlations. We discuss our approach in detail next.

\vspace{-0.1cm}
\subsection{Big Picture Policies (\bpp{})}
\label{subsec:bpp}
\vspace{-0.1cm}

To build intuition for our approach, consider the task of adding two spoons of sugar to a coffee. During execution, a robot may miss the spoon several times, producing observation histories that differ substantially from expert demonstrations, even though the task state remains the same. A policy conditioned on the full history must therefore implicitly learn to look for accurate information in a test-time history that is poorly covered by the data. \bpp{} instead conditions the policy on keyframes that correspond to behaviorally salient events, in this case the moment when sugar is successfully dropped into the coffee. With keyframes, trajectories that differ in incidental details collapse to a single, task-relevant representation. Thus, the policy is trained and evaluated on a much narrower and better-covered input distribution, despite variation in the raw history frames.

Concretely, \bpp{} first defines keyframes using concise, task-specific criteria that capture salient events. For example, in the password-entering tasks, these events are when a button is pressed. For a drawer search task, these events would be when a drawer is opened, and we can see the interior of the drawer.
The policy is then conditioned only on these keyframes, rather than the full history. Formally, let $\phi(o_t) \in \{0, 1\}$ denote a binary keyframe detector (e.g., a VLM) that identifies whether observation $o_t$ corresponds to a keyframe. To avoid redundant keyframes, we define the set of keyframes $\mathcal{K}$ as the timesteps corresponding to the onset of a detected event (a ``rising edge''):
\begin{align}
    \mathcal{K} = \{t : \phi(o_t) = 1 \land \phi(o_{t-1}) = 0\}.
    \label{eq:rising_edge}
\end{align}
\!\!This ensures that if consecutive detections occur, we only retain the first frame, and distinct events are separated by at least one negative detection.
Let $\mathcal{K}_t \subseteq \mathcal{K}$ denote the keyframes detected by time $t$. During policy learning, we account for the system latency, denoted by $\Delta$, inherent to the detection mechanism (e.g., 3 seconds for VLM queries). We define the \emph{latency-masked} keyframes available at time $t$ as:
\begin{align}
    \mathcal{K}_t^\Delta = \{k \in \mathcal{K}_t : k \le t - \Delta\}.
\end{align}
\!\!The policy $\pi_\theta(a_t|o_t, \{o_k\}_{k \in \mathcal{K}_t})$ conditions on the current frame and the available keyframes. For training, we use $\mathcal{K}_t^\Delta$ to simulate the latency at inference time, ensuring the policy learns to act under realistic delays. At inference, the policy simply conditions on all keyframes that have been detected.

\vspace{-0.1cm}
\subsection{Systems Design for \bpp}
\label{subsec:bpp_systems_design}
\vspace{-0.1cm}

For real-world tasks, we implement the keyframe detector $\phi$ using an off-the-shelf vision-language model (VLM) with a simple prompt as a binary classifier. Both for labeling demonstrations and during evaluation rollouts, we query the VLM at a frequency of 1 Hz using the current wrist camera image and the image from the previous query. Because keyframe definitions are often ambiguous in time, the rising edge condition in Eq.~\ref{eq:rising_edge} ensures that consecutive positive responses produce a single keyframe, avoiding duplicate detections. We use Gemini 3 Pro for detection, which incurs an average latency of 3 to 5 seconds per query, and set $\Delta = 3$ seconds during training to construct $\mathcal{K}_t^\Delta$. Processing fewer context frames also reduces training time. On $\mathrm{Drawer\ Search}$, \bpp{} reduces training time by 41\% compared to na\"ive history conditioning (see Appendix~\ref{sec:appendix_training_time}). 

\begin{AIbox}{Summary: \bpp{} alleviates the history-coverage problem.}
\begin{itemize}[leftmargin=*,itemsep=4pt,topsep=0pt]
    \item \bpp{} conditions on a small set of behaviorally salient events instead of uniformly subsampled histories, increasing the overlap between train and test inputs.
    \item \bpp{} detects keyframes by utilizing a vision-language model (VLM) and employs \emph{latency-masking} during training to more robustly handling delays at deployment time.
\end{itemize}
\end{AIbox}

\vspace{-0.2cm}
\section{Experiments}
\label{sec:experiments}
\vspace{-0.2cm}
We now evaluate \bpp{} on a suite of challenging manipulation tasks that require history conditioning. We first describe the real-world tasks used in our primary evaluation, then outline the baselines and methods we compare against, and finally analyze results across both real-world and simulation settings. Our experiments address four questions: \textbf{1)} Does keyframe-based history conditioning via BPP outperform methods with access to full histories on complex real-world tasks? \textbf{2)} When do representation auxiliary losses (e.g., \ptp{}) help or hurt? \textbf{3)} How does \bpp{} compare to prior approaches in data efficiency? \textbf{4)} How sensitive is \bpp{} to VLM detection errors? We answer these questions below.

\begin{figure*}[t]
    \centering
    \vspace{-0.4cm}
    \begin{tcolorbox}[
        width=1.00\textwidth,
        colback=figurebackground,
        boxrule=0pt,
        arc=10pt,
        left=6pt,
        right=6pt,
        top=4pt,
        bottom=4pt,
        boxsep=0pt,
        enhanced,
        colframe=figurebackground
    ]
    \centering
    \includegraphics[width=\textwidth]{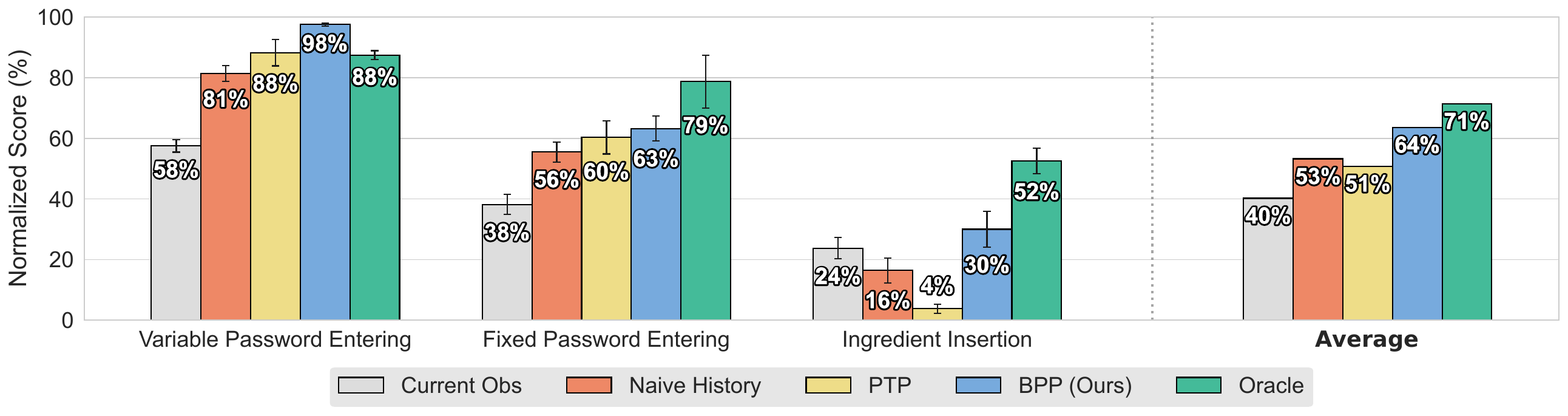}
    \end{tcolorbox}
    \vspace{-0.4cm}
    \caption{\textbf{Simulation results.} BPP outperforms all non-oracle methods, and even surpasses the oracle in one task, suggesting keyframe-based conditioning provides a representation more amenable to learning.}
    \label{fig:sim_combined_bar}
    \vspace{-0.4cm}
\end{figure*}

\vspace{-0.2cm}
\subsection{Real-World Benchmarks}
\label{subsec:real_world_benchmarks}
\vspace{-0.2cm}

We evaluate our approach on four real-world manipulation tasks using a bimanual ALOHA 2 platform~\cite{aldaco2024aloha}. The robot observes four RGB camera views (top, worm’s-eye, and two wrist-mounted) along with proprioceptive state, and outputs target joint positions and gripper commands for both arms. The control loop runs at 50 Hz for data collection and policy execution. All policies run locally on a workstation with an NVIDIA RTX 4090 GPU, while VLM inference for BPP is performed via the Vertex AI platform on Google Cloud. Data is collected via teleoperation by multiple operators with varying strategies and expertise to capture diverse demonstration quality. Appendix~\ref{sec:appendix_task_specs} provides detailed task specifications, including demonstrations, instructions, and success criteria.

\begin{itemize}[leftmargin=*,itemsep=3pt,topsep=-3pt]
    \item $\mathrm{Mug\ Replacement}$: Swap one of two mugs from a coffee machine for another (adapted from PTP~\cite{torne2025learning}). %
    \item $\mathrm{Marshmallows}$: Put two handfuls of marshmallows into a bowl, and then signal completion by pressing a red button.
    \item $\mathrm{Drawer\ Search}$: Search for a key in two cabinets with three drawers each, then place the key in the center of the table.
    \item $\mathrm{Stacking\ Puzzle}$: Given three different color pieces stacked in random order, sort them into locations of the same color and reassemble in the original configuration.
\end{itemize}
The difficulty of the $\mathrm{Mug\ Replacement}$ and $\mathrm{Marshmallows}$ tasks is comparable to the $\mathrm{Mug\ Swap}$ and $\mathrm{Two\ Scoops}$ tasks from prior work~\citep{torne2025learning}. A notable aspect of our data collection is that multiple teleoperators contributed demonstrations using different strategies, introducing additional behavioral diversity. In contrast, $\mathrm{Drawer\ Search}$ and $\mathrm{Stacking\ Puzzle}$ represent a significant increase in complexity. These long-horizon tasks require up to 90 seconds for successful execution and involve 4-5 sequential subtasks, demanding more complex history reasoning, such as tracking which drawers have been checked or recalling the initial puzzle configuration and task phase.

\vspace{-0.1cm}
\subsection{Baselines and Comparisons}
\label{subsec:methods}
\vspace{-0.1cm}

\begin{wraptable}{r}{0.6\linewidth}
    \centering
    \vspace{-0.3cm}
    \caption{\textbf{Real-Robot Experiments.} \bpp{} achieves nearly 70\% higher average success rate than the best prior approach (\ptp{}).}
    \label{tab:real_robot_results}
    \vspace{-0.2cm}
    \resizebox{0.99\linewidth}{!}{\begin{tabular}{lccc|c}
        \toprule
        Task & Current Obs & Na\"ive History & \ptp & \bpp{} \textbf{\textit{(Ours)}} \\
        \midrule
        Drawer Search & 11.1\% & 0.0\% & 0.0\% & \textbf{33.3\%} \\
        Marshmallows & 40.0\% & 25.0\% & 35.0\% & \textbf{65.0\%} \\
        Mug Replacement & 0.0\% & 5.0\% & 40.0\% & \textbf{60.0\%} \\
        Stacking Puzzle & 6.5\% & 21.0\% & 52.0\% & \textbf{56.0\%} \\
        \midrule
        Average & 14.4\% & 12.8\% & 31.8\% & \textbf{53.6\%} \\
        \bottomrule
    \end{tabular}}
    \vspace{-0.3cm}
\end{wraptable}
All comparisons use the same policy architecture (Diffusion Transformer) and training objective (DDPM~\citep{ho2020denoising} with action chunking of 50). We use separate image encoders per camera view, shared across all timesteps within each view. In addition to the \textbf{Current Observation} and \textbf{Na\"ive History} comparisons from Section~\ref{sec:analysis}, we compare \bpp{} to \textbf{Past-Token Prediction (\ptp{})}. \ptp{} uses the same inputs as na\"ive history conditioning but adds an auxiliary objective to predict actions at past timesteps. Unlike the official implementation of \ptp{}~\citep{torne2025learning}, we train the image encoder using gradients from all history steps. We also evaluate the \emph{Oracle} baseline, which has access to ground-truth state and thus requires no history conditioning (see Appendix~\ref{sec:appendix_sim_tasks}) for obtaining insights in simulation.

\vspace{-0.1cm}
\subsection{Real-World Results and Analysis}
\label{subsec:real_world_results}
\vspace{-0.1cm}

Our real-world results are summarized in Table~\ref{tab:real_robot_results}. BPP significantly outperforms all baselines, achieving nearly \textbf{70\% higher} performance than the best prior approach (PTP) on average. Beyond aggregate numbers, it is also very informative to analyze the behavioral differences between policies. We make the following observations in our experiments:

\textcolor{lightblue}{\emph{\textbf{(i) Policies that only utilize the current observation fail to track progress.}}} Without history, policies learn locally reasonable behaviors but cannot maintain task state. This manifests as repetitive loops (e.g. opening and closing the same drawer repeatedly, shuffling the same puzzle piece back and forth) and premature termination (e.g. pressing the completion button after a single handful of marshmallows instead of the required two). These failures are unsurprising: the tasks require history to succeed, and a memoryless policy has no mechanism to distinguish ``first handful completed'' from ``task complete''.

\textcolor{lightblue}{\emph{\textbf{(ii) Na\"ive history conditioning introduces new failure modes rather than solving old ones.}}} Unlike simulation results in Section~\ref{subsec:sim_results}, history conditioning \emph{hurts} performance relative to the current-observation baseline in two of four real-world tasks, with only modest gains in the others. We hypothesize that reversal in trends stems from additional sources of spurious correlations in the real-world data that are absent from simulation, such as background variation and sensor or actuator noise. The failure modes here differ qualitatively from memoryless policies, exhibiting degraded behaviors such as frequent failed grasps, stalling after opening drawers, and dropping objects outside target locations (Figures~\ref{fig:filmstrip_mug}, \ref{fig:filmstrip_drawer}, \ref{fig:filmstrip_stacking}). Most informatively, on $\mathrm{Marshmallows}$, a history-conditioned policy sometimes executes drop motions with an empty grasp or retries after a successful grasp, effectively replaying demonstration sequences regardless of the current observation. These behaviors indicate overfitting to training histories.

\begin{figure*}[t]
    \centering
    \vspace{-0.2cm}
    \includegraphics[width=0.95\textwidth]{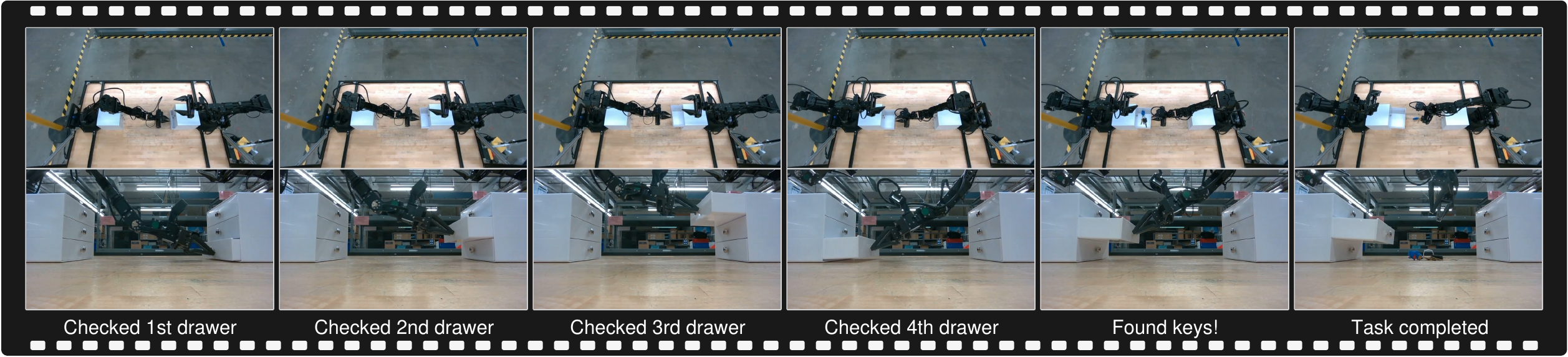}
    \vspace{-0.2cm}
    \caption{\textbf{Qualitative behavior on $\mathrm{Drawer\ Search}$.} \bpp{} systematically searches through drawers without revisiting them, demonstrating robust long-horizon progress tracking.}
    \label{fig:drawer_qualitative}
    \vspace{-0.4cm}
\end{figure*}

\begin{figure}[t]
    \centering
    \vspace{-0.4cm}
    \includegraphics[width=0.9\columnwidth]{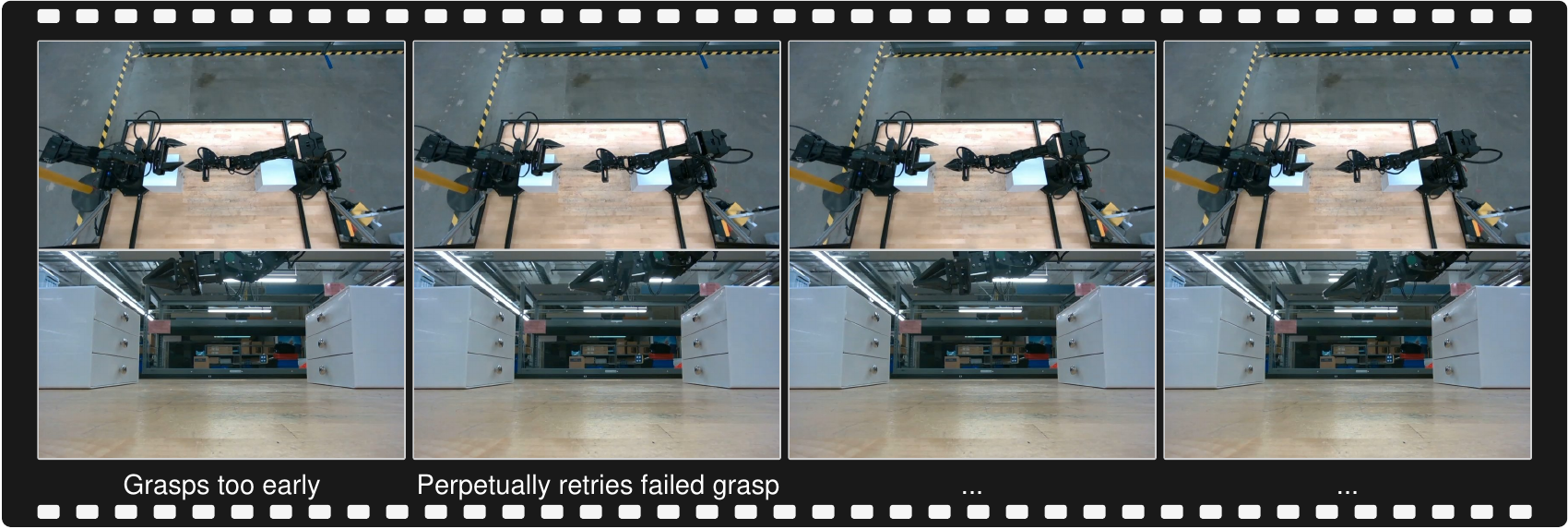}
    \caption{\textbf{\ptp{} failure on $\mathsf{Drawer\ Search}$.}
    \ptp{} fails to track progress after failed grasps or empty drawers, getting stuck in loops.}
    \label{fig:drawer_ptp_failure}
    \vspace{-0.4cm}
\end{figure}

\textcolor{lightblue}{\emph{\textbf{(iii) \ptp{} helps when predicting past actions helps infer the task state, but not otherwise.}}} \ptp{} shows significant improvements over the na\"ive history conditioning approach on $\mathrm{Mug\ Replacement}$ and $\mathrm{Stacking\ Puzzle}$, but offers little benefit on $\mathrm{Drawer\ Search}$ and $\mathrm{Marshmallows}$. As we shall see later, this pattern of task-dependent performance is repeated in our simulated benchmarks. To explain this pattern, let us consider what the auxiliary loss actually requires. In $\mathrm{Stacking\ Puzzle}$, correctly predicting past joint positions requires knowing which pieces have been placed and whether the policy is in the unstacking or restacking phase, which are also essential for correctly choosing subsequent actions. Similarly in $\mathrm{Mug\ Replacement}$, predicting past actions requires knowing whether each mug has been placed in the machine or returned to its plate. However, \emph{this heuristic breaks down when past actions do not reliably indicate task state}. In $\mathrm{Drawer\ Search}$, the robot may pull away from a drawer either after discovering no key \textit{or} after a failed grasp attempt (see Fig.~\ref{fig:drawer_ptp_failure}). Similarly, in $\mathrm{Marshmallows}$, pulling away from the pile (expressed as target joint positions) appears identical whether or not the gripper successfully grasped any marshmallows. In both cases, the same action sequence is consistent with multiple task states, so the auxiliary prediction no longer allows task state disambiguation. In this scenario, \ptp{} may not offer representational benefits. We observe similar results in simulation (Section~\ref{subsec:sim_results}).

\textcolor{lightblue}{\emph{\textbf{(iv) \bpp{} produces robust progress tracking.}}}
\bpp{} performs best across all four tasks by conditioning on a compact, semantically meaningful history summary rather than raw observations. Behaviorally, keyframe conditioning gives the policy a clearer sense of task progress: it retries more reliably after failed grasps and, in $\mathrm{Drawer\ Search}$, executes systematic searches that avoid revisiting checked drawers (Fig.~\ref{fig:drawer_qualitative}). \bpp{} failures fall into three categories. Most failures stem from data limitations: in $\mathrm{Drawer\ Search}$, failures arise from dropping keys inside drawers, which never appears in training, or repeated drawer-opening failures. Rarer errors come from VLM false positives, for example misclassifying an empty-handed drop as a successful scoop in $\mathrm{Marshmallows}$, leading to a premature button press. Finally, some errors are due to VLM latency. While usually irrelevant, in $\mathrm{Mug\ Replacement}$ trials where the top mug starts in the machine, the grasp event and subsequent decision occur close enough in time that delayed keyframe detection degrades performance.

\vspace{-0.2cm}
\subsection{Analysis in Simulation}
\label{subsec:sim_results}
\vspace{-0.2cm}

Next, we compare our approach to the \textbf{Oracle} baseline approach, where the policy has access to privileged ground-truth state information in simulation in addition to the current observation. We evaluate this on the simulation tasks from Section~\ref{sec:analysis}. For instance, in $\mathrm{Ingredient\text{-}Insertion}$, the policy receives a one-hot vector indicating the number of lemons already inserted (see Appendix~\ref{sec:appendix_sim_tasks}). This comparison estimates the achievable performance if challenges in inferring and reasoning over history were fully resolved. Figure~\ref{fig:sim_combined_bar} summarizes the results and leads to several key observations.

\textcolor{lightblue}{\emph{\textbf{(i) PTP performs comparably to Na\"ive History in simulation}}}, except on $\mathrm{Ingredient\text{-}Insertion}$. In Section~\ref{subsec:real_world_results}, we argued that alignment between past-action prediction and task state inference determined PTP's effectiveness. Here, we apply the same argument. In the password-entering tasks, predicting target joint positions for each past observation is well-aligned with the task state, as it requires knowing which buttons were pressed and in which order. In $\mathrm{Ingredient\text{-}Insertion}$, the relevant task state is simple: how many lemons have been inserted so far. Predicting past actions does encode this information, but it also encodes the specific spatial locations of the lemons that the demonstrator happened to pick. Since lemon positions were randomized across episodes and demonstrators chose arbitrarily, predicting these specific locations will lead to overfitting. Finally, it may seem contradictory that explicit ground-truth state prediction hurts performance (Section~\ref{sec:analysis}) while PTP does not, since predicting past tokens should impose a similar regularization. However, PTP does not explicitly optimize a state prediction loss; instead, it applies a weaker signal of past token prediction, which may only encourage useful structure in the learned features without causing them to overfit to the training distribution.

\textcolor{lightblue}{\emph{\textbf{(ii) BPP performs best; can sometimes outperform oracle.}}} BPP significantly outperforms all non-oracle comparisons across all tasks. In addition, BPP achieves a higher performance than \textit{Oracle} on $\mathrm{Variable\text{-}Password}$. We attribute this surprising result to the possibility that BPP provides a representation of history that is more amenable to learning a generalizable policy than the oracle one-hot encoding.

\vspace{-0.2cm}
\subsection{Ablation Studies in Simulation}
\label{subsec:ablations}
\vspace{-0.2cm}

To further understand the sources of the performance gap between methods, we conduct two complementary ablation experiments.

\begin{wrapfigure}{r}{0.45\textwidth}
    \vspace{-0.6cm}
    \begin{tcolorbox}[
        colback=figurebackground,
        boxrule=0pt,
        arc=10pt,
        left=6pt,
        right=6pt,
        top=4pt,
        bottom=4pt,
        boxsep=0pt,
        enhanced,
        colframe=figurebackground
    ]
    \centering
    \includegraphics[width=\linewidth]{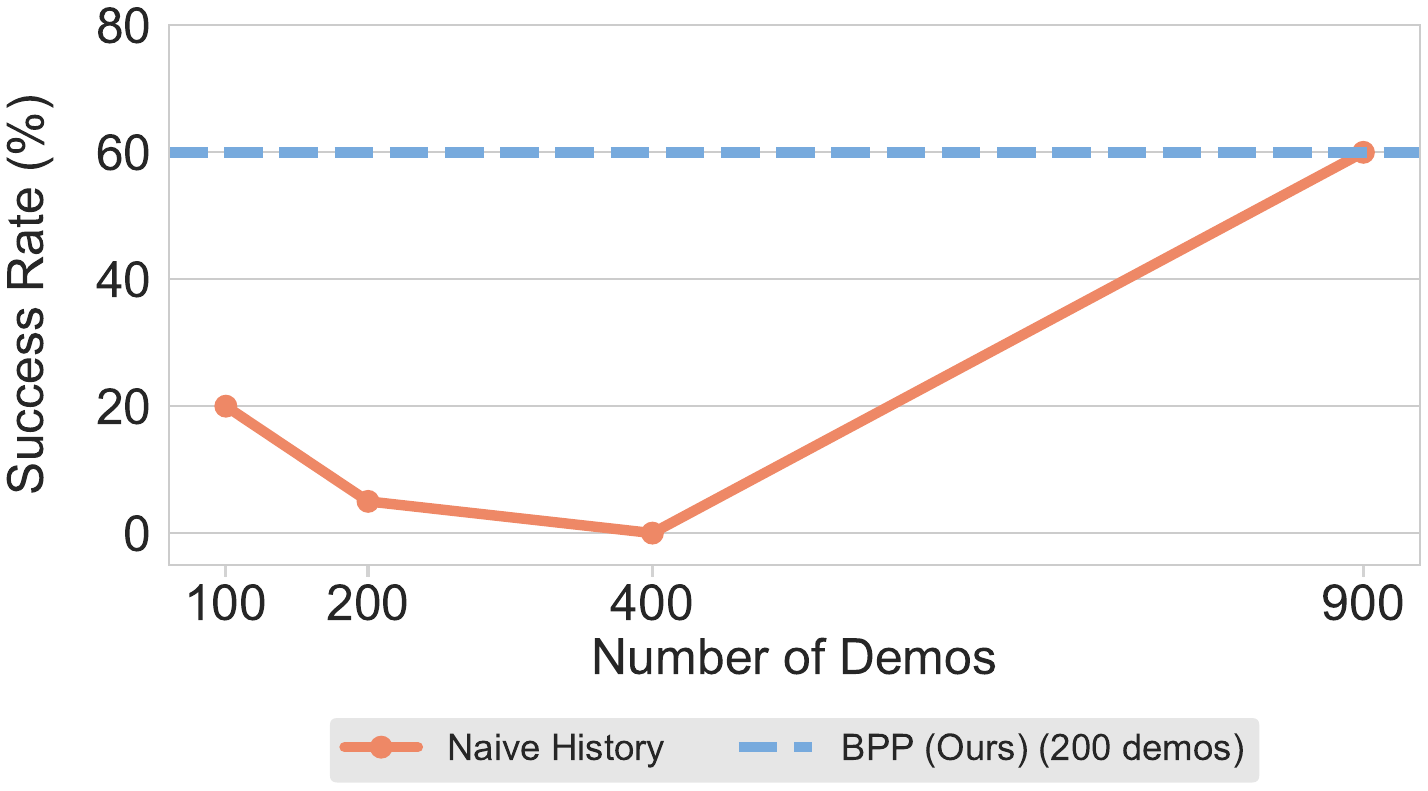}
    \end{tcolorbox}
    \vspace{-0.4cm}
    \caption{\textbf{Data efficiency on the Mug task.} With sufficient demonstrations, na\"ive history conditioning catches up with BPP.}
    \label{fig:num_demos_ablation}
    \vspace{-0.5cm}
\end{wrapfigure}
\textbf{Data efficiency.} If na\"ively conditioning on observation history underperforms by latching onto spurious temporal correlations, then providing more demonstrations should eventually allow the model to discover structures that generalize. We test this hypothesis on $\mathrm{Mug\ Replacement}$ by training with varying dataset sizes (Figure~\ref{fig:num_demos_ablation}). Performance evolves non-monotonically: adding demonstrations initially hurts Na\"ive History, likely due to the increased diversity of behaviors that expose more failure modes. However, with enough data, Na\"ive History matches the success rate of BPP. BPP achieves strong performance with far less data by reducing exposure to spurious history cues.

\begin{wraptable}{r}{0.4\linewidth}
    \centering
    \vspace{-0.3cm}
    \resizebox{0.99\linewidth}{!}{\begin{tabular}{lc}
        \toprule
        Method & Success Rate (\%) \\
        \midrule
        BPP (w/ VLM keyframes) & 60\% \\
        BPP w/ Oracle keyframes & 70\% \\
        Na\"ive History & 5\% \\
        PTP & 40\% \\
        \bottomrule
    \end{tabular}}
    \vspace{-0.2cm}
    \caption{VLM-based keyframe labeling incurs only a modest performance drop compared to oracle keyframes, demonstrating that \bpp{} is robust to imperfect VLM detection. Even with VLM inference latency, \bpp{} significantly outperforms prior history-conditioning approaches.}
    \label{tab:mug_comparison}
    \vspace{-0.4cm}
\end{wraptable}
\textbf{VLM keyframes versus ground-truth keyframes.} We also quantify how much performance drops due to imperfect keyframe detection. For this comparison, we construct an ``oracle'' keyframe baseline on the $\mathrm{Mug\ Replacement}$ task by supplying the first observation of each episode as an additional context frame, which contains enough information to disambiguate the task. This oracle variant achieves a 70\% success rate, compared to 60\% with VLM-selected keyframes, indicating that keyframe recognition errors explain only a small portion of the performance gap (Table~\ref{tab:mug_comparison}).

\vspace{-0.2cm}
\section{Discussion and Conclusion}
\vspace{-0.2cm}

We show that the primary failure mode of history-conditioned imitation learning stems not from architectural limits, but from insufficient coverage over the exponentially large space of possible histories. To address this, we introduce Big Picture Policies (\bpp{}), which use off-the-shelf vision-language models to compress observation histories into a compact sequence of task-relevant keyframes. By conditioning policies on these semantic events rather than raw temporal sequences, BPP reduces distribution shift between training and deployment. Across three simulation and four real-world tasks, \bpp{} enables robust progress tracking and outperforms state-of-the-art history-conditioned approaches by up to 70\%. 

\textbf{Limitations.} \bpp{} relies on the inference speed of an external VLM; while we mitigate this during training via latency masking, real-time deployment remains constrained by VLM query latency, which can be critical for highly dynamic tasks (e.g., parts of Mug Replacement). \bpp{} also relies on accurate keyframe detection. Although our ablations show robustness to some errors, systematic false positives, such as misclassifying an empty grasp, can trigger premature state transitions. Future work could address these issues by distilling keyframe detection into low-latency models. More broadly, BPP points toward event-based robot learning, where abstracting continuous control into semantic milestones offers a promising path for scaling to multi-stage, long-horizon tasks using diverse, non-expert data.

\textbf{Future Directions.} While BPP currently relies on task-specific VLM prompting, the salient event definitions used for keyframe detection are simple enough that large language models could feasibly generate them at scale for new tasks. This opens a path toward automatically labeling keyframes across large, diverse datasets -- a key ingredient for training history-conditioned generalist robot policies. Beyond tasks that explicitly require memory, the core insight of BPP, that conditioning on semantically meaningful keyframes rather than raw observation histories yields more robust and efficient learning, extends naturally to other settings. In particular, tasks that demand fast adaptation or forms of in-context learning stand to benefit from similar abstractions. Consider, for example, a policy that must learn from past failed attempts: a single point observation (i.e., an image and state from one timestep) may not capture enough information to understand why a grasp failed. In such cases, the notion of a keyframe can be generalized to \emph{key segments}, or short sequences of consecutive frames that capture the relevant event, such as a failed grasp, providing richer context for downstream learning. Extending \bpp{} to these settings is an important direction for future work.

\vspace{-0.2cm}
\section*{Acknowledgements}
\vspace{-0.2cm}
We thank the robot operators from GDM Robotics for their efforts in collecting demonstrations and conducting policy evaluations. We are also grateful to the members of the CMU AIRe lab and GDM Robotics for their valuable discussions and feedback on this work. This work was done when MSM was a student researcher at Google DeepMind.

\bibliography{references}

\newpage
\appendix
\onecolumn
\part*{Appendices}

\raggedbottom
\section{Additional Results}
\label{sec:additional_results}

\subsection{Frozen Encoder Ablation}
\label{sec:appendix_frozen_encoder}
To investigate whether the image encoder needs to adapt when processing historical observations, we compare na\"ive history conditioning with a variant where the image encoder is frozen and initialized from a policy trained on current observations only. As shown in Table~\ref{tab:frozen_encoder_ablation}, freezing the encoder leads to a significant drop in performance, indicating that the encoder must learn to extract different features from historical frames compared to the current observation.

\begin{table}[h]
    \centering
    \begin{tabular}{lc}
        \toprule
        Method & Success Rate (\%) \\
        \midrule
        Na\"ive History & $55.5 \pm 3.3$ \\
        Na\"ive History + Frozen Encoder & $43.5 \pm 2.2$ \\
        \bottomrule
    \end{tabular}
    \caption{Frozen encoder ablation on the Fixed Password Entering task. Freezing the image encoder (initialized from a current-observation-only policy) reduces performance.}
    \label{tab:frozen_encoder_ablation}
\end{table}

\vspace{1em}
\section{Implementation Details}
\label{sec:appendix_details}

\subsection{Policy Architecture}
\label{sec:appendix_policy_architecture}

\begin{figure}[h!]
    \centering
    \includegraphics[width=\textwidth]{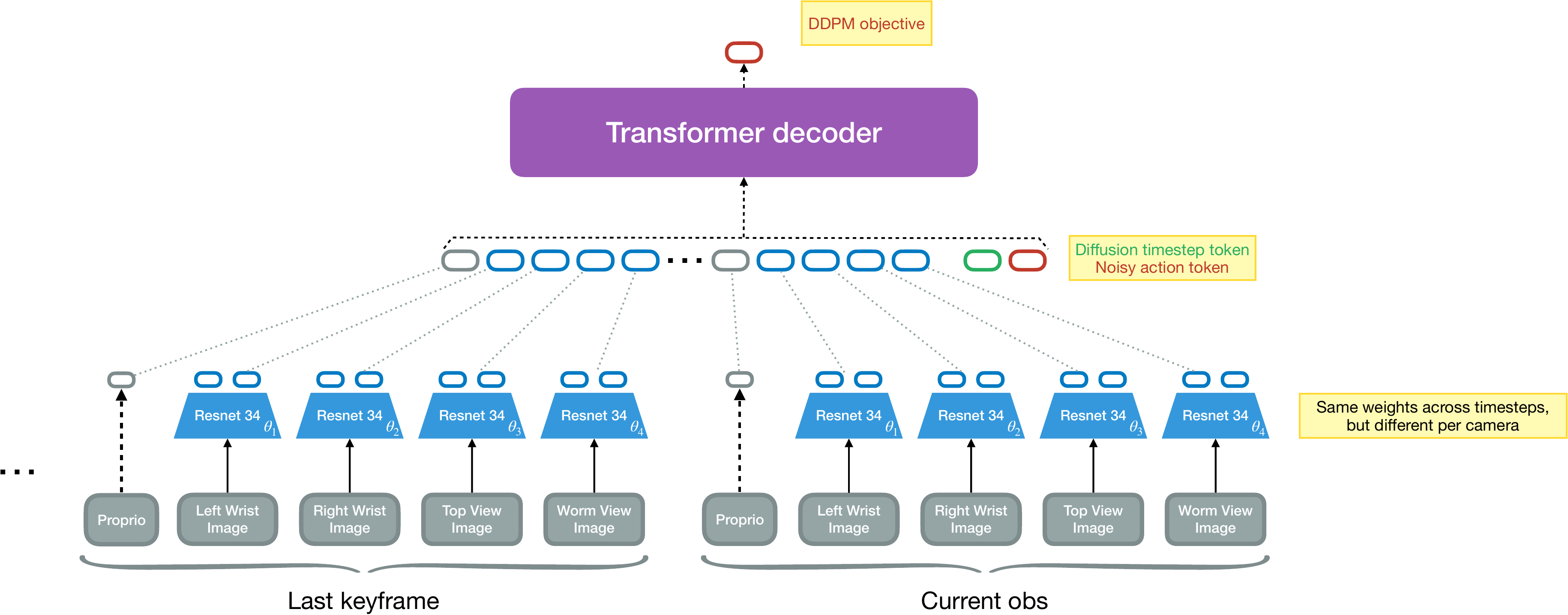}
    \caption{Policy Architecture. We use a ResNet34 to encode images from 4 cameras, and a transformer decoder to denoise actions.}
    \label{fig:policy_architecture}
\end{figure}

We visualize the policy architecture in Figure~\ref{fig:policy_architecture}.
Each timestep consists of 4 camera views (top view, worms-eye view, and two wrist-mounted cameras). We use a standard ResNet34 image encoder, with different weights for each camera view, but shared across timesteps. We flatten the resulting image features into tokens. Additionally, we project the proprioceptive state into a token, and use a learnable token for the diffusion timestep and another for the action to denoise.
We then pass all these tokens through a Transformer decoder consisting of 7 layers with layer size 512, 8 attention heads, and a dropout rate of 0.1.
The policy outputs action chunks of 50 timesteps (unless otherwise specified). The training objective is DDPM~\citep{ho2020denoising}.

\newpage
\subsection{Training Time Comparison}
\label{sec:appendix_training_time}

Table~\ref{tab:drawer_training_time} compares the training time of different methods on the $\mathrm{Drawer\ Search}$ task.

\begin{table}[h]
    \centering
    \begin{tabular}{lc}
        \toprule
        Method & Training Time (hours) \\
        \midrule
        Current Obs Only & 9.95 \\
        Na\"ive History & 21.67 (\textbf{+118\%}) \\
        BPP & 12.84 (\textbf{+29\%}) \\
        \bottomrule
    \end{tabular}
    \caption{Training time comparison on $\mathrm{Drawer\ Search}$. Na\"ive history conditioning significantly increases training time, while BPP adds only modest overhead relative to training on current observations only.}
    \label{tab:drawer_training_time}
\end{table}

\subsection{History Parameters}
\label{sec:appendix_task_details}
\begin{table}[h]
    \centering
    \footnotesize
    \begin{tabular}{l c c c}
        \toprule
        \textbf{Task Name} & \textbf{Obs. Horizon} & \textbf{Stride} & \textbf{\# Past Frames} \\
        \midrule
        \multicolumn{4}{l}{\textbf{Real-World Tasks}} \\
        \hspace{3mm}$\mathrm{Mug\ Replacement}$ & 12~s & 3~s & 4 \\
        \hspace{3mm}$\mathrm{Marshmallows}$ & 12~s & 3~s & 4 \\
        \hspace{3mm}$\mathrm{Drawer\ Search}$ & 25~s & 5~s & 5 \\
        \hspace{3mm}$\mathrm{Stacking\ Puzzle}$ & 40~s & 8~s & 5 \\
        \midrule
        \multicolumn{4}{l}{\textbf{Simulation Tasks}} \\
        \hspace{3mm}$\mathrm{Ingredient\ Insertion}$ & 13~s & 2.6~s & 5 \\
        \hspace{3mm}$\mathrm{Fixed\ Password}$ & 4~s & 1~s & 4 \\
        \hspace{3mm}$\mathrm{Variable\ Password}$ & 4~s & 1~s & 4 \\
        \bottomrule
    \end{tabular}
    \caption{History parameters for Na\"ive History Conditioning and PTP.}
    \label{tab:history_parameters}
\end{table}

\vspace{1em}
\section{Task Specifications}
\label{sec:appendix_task_specs}

\subsection{Real-World Tasks}

\medskip
\noindent \textbf{Mug Replacement}
\begin{itemize}[leftmargin=*]
    \item \textbf{Task Description:} The robot must swap one of two mugs from a coffee machine for another.
    \item \textbf{Success Conditions:} The mug initially in the coffee machine must be placed upright inside the plate that started empty, and the other mug must be placed fully inside the coffee machine.
    \item \textbf{Data Collection:} We collected a total of 900 demonstrations. The main evaluation uses 200 of these demos (the same subset for all methods), and the rest are used for the data efficiency ablation (Section~\ref{subsec:ablations}). Data collection was performed by 5 different operators, each with naturally varying strategies and teleoperation speeds.
    \item \textbf{Initial Conditions:} The coffee machine and plates were fixed across episodes, but the top mug started in the coffee machine in approximately 50\% of the demonstrations, while the bottom mug started in the machine for the other 50\%. We also randomized the rotation of the mugs both on the plates and in the coffee machine.
    \item \textbf{Key Frames:} Defined as moments in which the robot hand had just picked up a mug, yielding a maximum of 2 key frames per episode. The VLM prompt used for this task is detailed in Appendix~\ref{sec:appendix_vlm_prompts}.
\end{itemize}

\medskip
\noindent \textbf{Marshmallows}
\begin{itemize}[leftmargin=*]
    \item \textbf{Task Description:} The robot must pick and place two handfuls (at least one marshmallow per handful) from the large bowl into the smaller red bowl, and subsequently press the red button to signal completion.
    \item \textbf{Success Conditions:} A handful transfer is considered successful if at least one marshmallow enters and remains in the red bowl. The entire episode is successful only if exactly two handfuls are successfully transferred, followed by a button press.
    \item \textbf{Data Collection:} We collected 250 demonstrations using a single teleoperator. The dataset includes natural retries, as a portion of the grasp attempts were unsuccessful during collection.
    \item \textbf{Initial Conditions:} The positions of the bowls and the red button were fixed. Crucially, the smaller target bowl was emptied into the source bowl only every 2--3 episodes. This variation ensures that the task requires history conditioning; if the bowl always started empty, a policy could simply stop after placing a handful whenever it detected marshmallows in the target bowl. During evaluation, the target bowl always started with 3 marshmallows to test this ambiguity.
    \item \textbf{Key Frames:} Defined as moments in which marshmallows are dropped into the red bowl. The VLM prompt used for this task is detailed in Appendix~\ref{sec:appendix_vlm_prompts}.
\end{itemize}

\medskip
\noindent \textbf{Drawer Search}
\begin{itemize}[leftmargin=*]
    \item \textbf{Task Description:} There are two cabinets with three drawers each. One of the six drawers has a set of keys hidden inside. The task is to find the keys, and drop them in the middle of the table.
    \item \textbf{Success Conditions:} An episode is successful whenever the keys touch the middle of the table.
    \item \textbf{Data Collection:} We collected 200 demonstrations using a single teleoperator. For the purposes of testing history conditioning for search tasks, we used a non-markovian data collection strategy: Before starting the episode, the data collector decides whether to start with the left cabinet or right cabinet, top to bottom or bottom to top. If the keys are not found in the first cabinet, the data collector decides again whether to search the second cabinet top to bottom or bottom to top.
    \item \textbf{Initial Conditions:} During evaluations, we test each of the six initial key positions three times, for a total of 18 evaluations.
    \item \textbf{Key Frames:} Defined as moments in which a drawer is newly opened. Note that we prompt the VLM to simply detect whether a drawer is open, and we rely on the rising edge logic to only keep newly opened drawers. The VLM prompt used for this task is detailed in Appendix~\ref{sec:appendix_vlm_prompts}.
\end{itemize}

\medskip
\noindent \textbf{Stacking Puzzle}
\begin{itemize}[leftmargin=*]
    \item \textbf{Task Description:} Three square pieces are stacked in the middle of the table in random order. The task is to take out piece by piece and put them on top of the circle of the same color on the right of the table; when there are no pieces left, the robot must put them back in the original order.
    \item \textbf{Success Conditions:} Since this task is very long-horizon and total success rates are very low (2 out of 18 for BPP and PTP, 0 for other baselines), we report ``score'', which is the average percentage of task completion. For each piece put in the right place, we add 1/6 of the score. A piece counts as being correctly placed if it ends up touching the correct circle (if in the first phase of movement), or if it is stacked in the right order (in the second phase). Rollouts are terminated if a piece is placed in the wrong place.
    \item \textbf{Data Collection:} Three people collected 200 demonstrations. The ordering of pieces was randomized during data collection.
    \item \textbf{Initial Conditions:} During evaluations, we test each of the six orderings of pieces three times, for a total of 18 evaluations.
    \item \textbf{Key Frames:} Defined as moments in which a piece is picked up. While episodes have 6 moments in which pieces are picked up, we only use the first 3 for efficiency, since they are sufficient for task understanding. The VLM prompt used for this task is detailed in Appendix~\ref{sec:appendix_vlm_prompts}.
\end{itemize}

\subsection{Simulation Tasks}
\label{sec:appendix_sim_tasks}

We used a Meta Quest 3 for teleoperation and bimanual control. We found that substantially more data was needed in simulation to achieve reasonable behaviors than in the real world, likely due to lower quality of data and increased behavior diversity.

\medskip
\noindent \textbf{Ingredient-Insertion}
\begin{itemize}[leftmargin=*]
    \item \textbf{Task Description:} The task is to insert two lemons into the bowl, and then press the button. This task is meant to emulate the task of inserting two spoons of sugar in coffee, but simplified for simulation. Crucially, when lemons touch the bowl, they become invisible, such that the policy cannot simply count the number of lemons inside the bowl to determine the state. Additionally, the number and position of lemons on the table are randomized, so the policy cannot simply count the visible lemons either.
    \item \textbf{Success Conditions:} An episode is successful only if exactly two lemons have been dropped in the bowl, and then the button is pressed.
    \item \textbf{Data Collection:} We collected 900 demonstrations using three different teleoperators.
    \item \textbf{Initial Conditions:} The number of lemons on the table was randomized uniformly between 6 and 10, and they were placed randomly on the table.
    \item \textbf{Key Frames:} Defined as moments in which the robot lets go of a lemon. As such, we condition on two keyframes per episode.
    \item \textbf{Oracle history-state information:} For the Oracle baseline, we grant access to a one-hot encoding of how many lemons have been placed in the bowl so far (0--2). For the ground-truth state prediction experiment, we predict this same one-hot encoding, formulating it as a three-way classification problem.
\end{itemize}

\medskip
\noindent \textbf{Fixed-Password}
\begin{itemize}[leftmargin=*]
    \item \textbf{Task Description:} The robot has to input the password ``1112134'', which has 7 digits. We selected this password because it requires variable amounts of memory at different stages of the task. For example, merely remembering the last digit entered is insufficient; if the last digit entered was 1, the next could be a 1, a 2, or a 3. The buttons are arranged in a $3 \times 3$ grid, and the ordering of the buttons is randomized in each trial.
    \item \textbf{Success Conditions:} For both password tasks, we report the percentage of buttons pressed in the correct order (calculated as $(7 - \text{Hamming distance}) / 7$). Trajectories are terminated when the robot presses 7 numbers.
    \item \textbf{Data Collection:} We collected 600 demonstrations from two teleoperators.
    \item \textbf{Initial Conditions:} The order of the numbers on the buttons is randomized for each episode.
    \item \textbf{Key Frames:} Defined as the moment in which the robot touches a button.
    \item \textbf{Oracle history-state information:} We provide a one-hot encoding of the number of buttons pressed so far (0--6). For the ground-truth state prediction experiment, we predict this same one-hot encoding, formulating it as a seven-way classification problem.
\end{itemize}

\medskip
\noindent \textbf{Variable-Password}
\begin{itemize}[leftmargin=*]
    \item \textbf{Task Description:} There are 3 buttons (1--3), and a target password of 4 digits (such that there is at least one repeated digit in the password). The task is to insert the password correctly. The policy receives the target password as a matrix of 4 one-hot encodings.
    \item \textbf{Success Conditions:} As in the fixed password task, we report the percentage of buttons pressed in the correct order (calculated as $(4 - \text{Hamming distance}) / 4$). Trajectories are terminated when the robot presses 4 numbers.
    \item \textbf{Data Collection:} We collected 900 demonstrations using 4 teleoperators.
    \item \textbf{Initial Conditions:} The order of the buttons is always the same, although the target password is randomized for each episode.
    \item \textbf{Key Frames:} Defined as the moment in which the robot touches a button.
    \item \textbf{Oracle history-state information:} We provide a one-hot encoding of the number of buttons pressed so far (0--3).
\end{itemize}

\vspace{1em}
\section{Real-World Task and Evaluation Details}
\label{sec:appendix_real_world}

\subsection{VLM Prompts for Keyframe Selection}
\label{sec:appendix_vlm_prompts}

We provide the VLM prompts used for keyframe selection in our real-world tasks below.

\begin{tcolorbox}[title=Mug Replacement Keyframe Prompt, colback=white, colframe=black, fonttitle=\bfseries]
\textit{Decide whether or not the robot hand has just picked up a mug. The PDF attached which says Current observation to label and Previous observation number 1, contains both the current observation and the immediate prior observation. If the robot gripper is completely grabbing a mug (no space between the gripper and the mug), and it wasn't grabbing it in the previous observation, then answer YES. Otherwise, answer NO. Only output the final answer, no justification.}
\end{tcolorbox}

\begin{tcolorbox}[title=Marshmallows Keyframe Prompt, colback=white, colframe=black, fonttitle=\bfseries]
\textit{Decide whether or not the robot hand has just dropped marshmallows into the red bowl. The PDF attached which says Current observation to label and Previous observation number 1, contains both the current observation and the immediate prior observation. If the robot gripper is open and on top of the red bowl, and it was closed before, then answer YES. Otherwise, answer NO. Only output the final answer, no justification.}
\end{tcolorbox}

\begin{tcolorbox}[title=Drawer Search Keyframe Prompt, colback=white, colframe=black, fonttitle=\bfseries]
\textit{The user will attach a PDF with wrist-mounted camera images from a robot that has to open drawers. Your task is to determine whether in any of the two wrists, a drawer is open (only counting the small white cabinets, not anything on the background). If in any of the two views a drawer is open, answer YES. Otherwise, answer NO. Only output the final answer, no justification.}
\end{tcolorbox}

\begin{tcolorbox}[title=Stacking Puzzle Keyframe Prompt, colback=white, colframe=black, fonttitle=\bfseries]
\textit{Decide whether or not the robot hand has just picked up a piece. The PDF attached which says Current observation to label and Previous observation number 1, contains both the current observation and the immediate prior observation. If the robot gripper is completely grabbing a piece (no space between the gripper and the piece), and it wasn't grabbing it in the previous observation, then answer YES. Otherwise, answer NO. Only output the final answer, no justification.}
\end{tcolorbox}

\newsavebox{\filmstripbox}
\newlength{\filmstripheight}
\savebox{\filmstripbox}{\includegraphics[width=\textwidth]{figures/filmstrips/drawer_keyframes_success_long_horizon_filmstrip/drawer_keyframes_success_long_horizon_filmstrip_smaller-crop.pdf}}
\setlength{\filmstripheight}{0.8\ht\filmstripbox}

\clearpage
\section{Evaluation Filmstrips}
\label{sec:appendix_filmstrips}

\noindent In this section, we provide visualization of evaluation episodes across our real-world tasks using ``filmstrips'' that capture sequences of frames. All filmstrips in the following figures are scaled to have the same height, with the longest sequence in each task set spanning the full width of the page.

\subsection{$\mathrm{Drawer\ Search}$}
\begin{figure}[H]
    \centering
    \begin{tcolorbox}[
        width=0.95\textwidth,
        center,
        colback=figurebackground,
        boxrule=0pt,
        arc=10pt,
        left=6pt,
        right=6pt,
        top=4pt,
        bottom=4pt,
        boxsep=0pt,
        enhanced,
        colframe=figurebackground
    ]
    \centering
    \footnotesize\textbf{BPP (Success)} \\[1pt]
    \includegraphics[height=\filmstripheight]{figures/filmstrips/drawer_keyframes_success_long_horizon_filmstrip/drawer_keyframes_success_long_horizon_filmstrip_smaller-crop.pdf} \\[6pt]
    \footnotesize\textbf{PTP (Failure)} \\[1pt]
    \includegraphics[height=\filmstripheight]{figures/filmstrips/drawer_ptp_failure_filmstrip/drawer_ptp_failure_filmstrip_smaller-crop.pdf} \\[6pt]
    \footnotesize\textbf{Na\"ive History (Failure)} \\[1pt]
    \includegraphics[height=\filmstripheight]{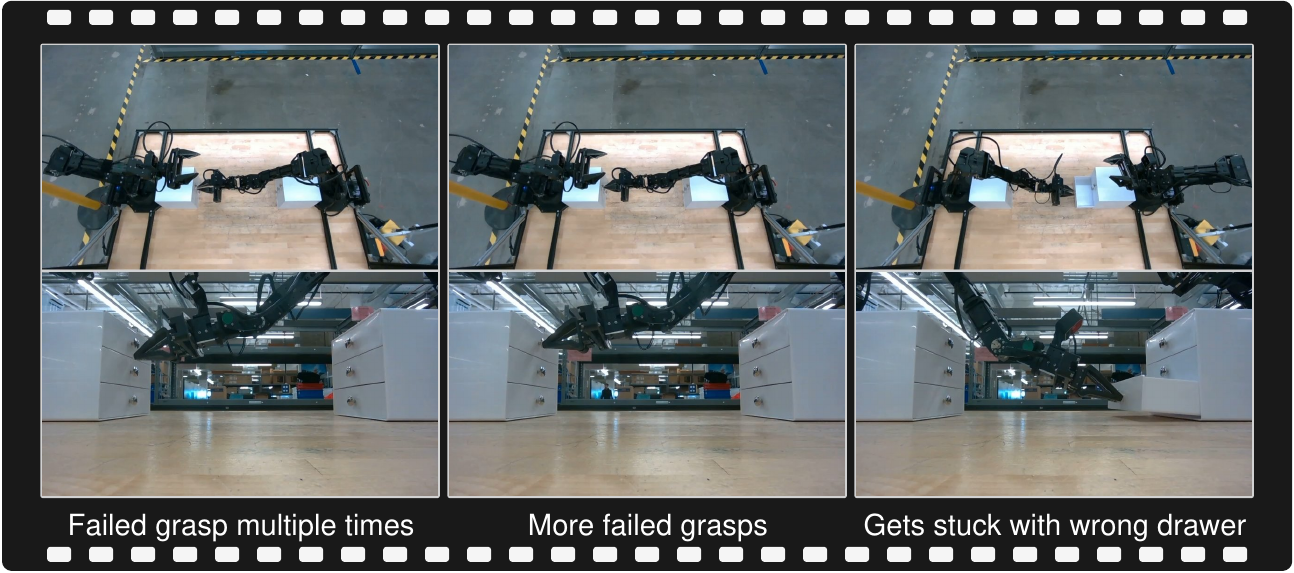} \\[6pt]
    \footnotesize\textbf{Current Obs (Failure)} \\[1pt]
    \includegraphics[height=\filmstripheight]{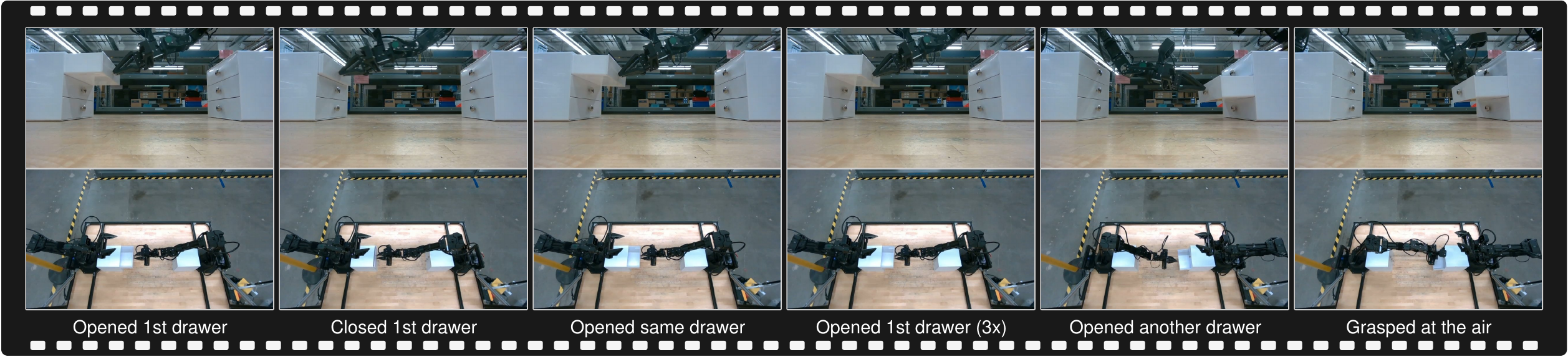}
    \end{tcolorbox}
    \caption{\textbf{$\mathrm{Drawer\ Search}$ evaluation behaviors.} Our method successfully maintains long-term context to search multiple drawers, while baselines fail to progress through the sequence.}
    \label{fig:filmstrip_drawer}
\end{figure}

\clearpage
\subsection{Marshmallow}
\begin{figure}[H]
    \centering
    \begin{tcolorbox}[
        width=0.95\textwidth,
        center,
        colback=figurebackground,
        boxrule=0pt,
        arc=10pt,
        left=6pt,
        right=6pt,
        top=4pt,
        bottom=4pt,
        boxsep=0pt,
        enhanced,
        colframe=figurebackground
    ]
    \centering
    \footnotesize\textbf{BPP (Success)} \\[1pt]
    \includegraphics[height=\filmstripheight]{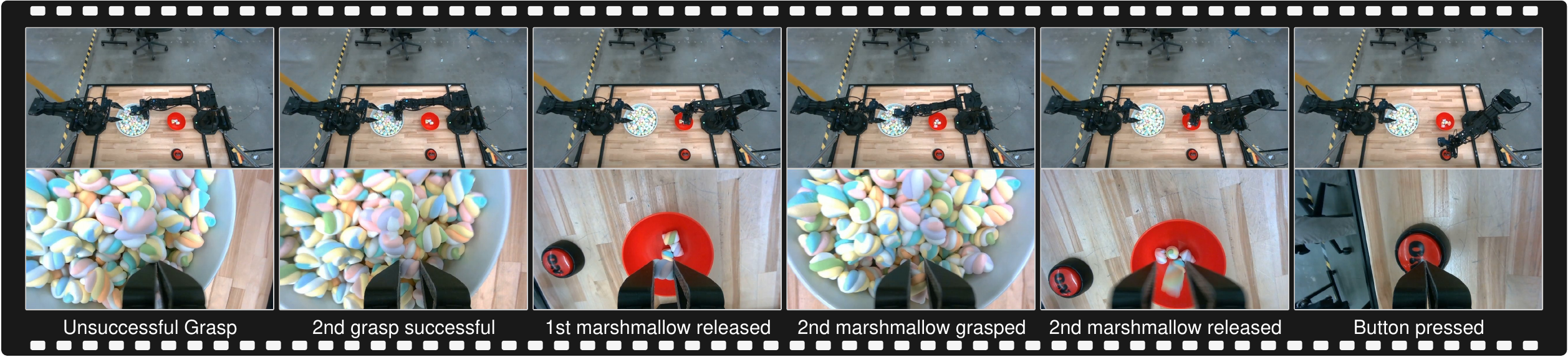} \\[6pt]
    \footnotesize\textbf{BPP (Failure)} \\[1pt]
    \includegraphics[height=\filmstripheight]{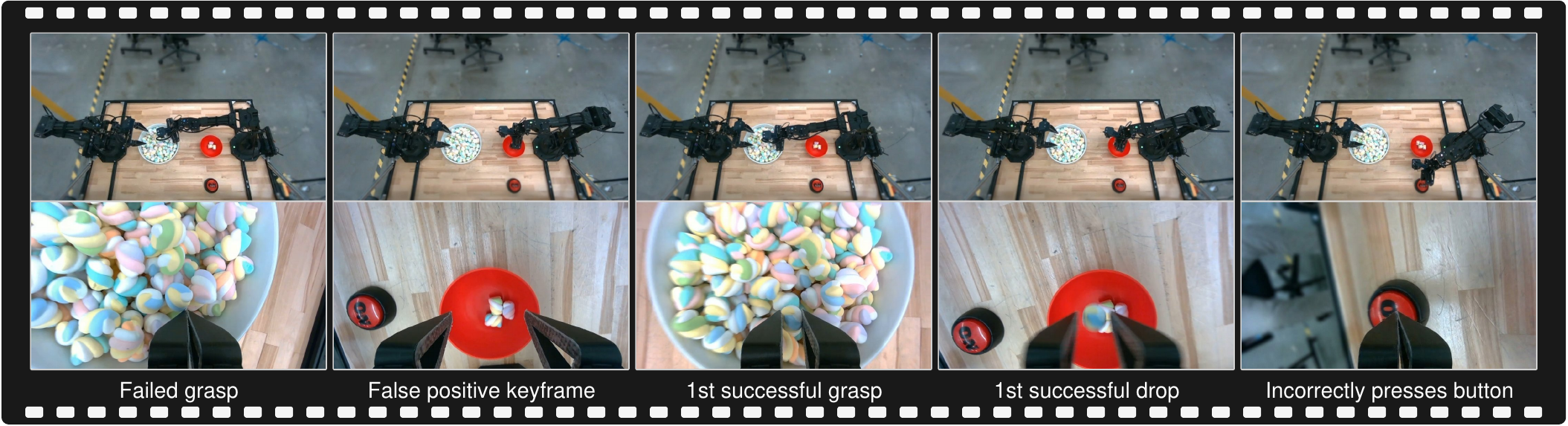} \\[6pt]
    \footnotesize\textbf{Na\"ive History (Failure)} \\[1pt]
    \includegraphics[height=\filmstripheight]{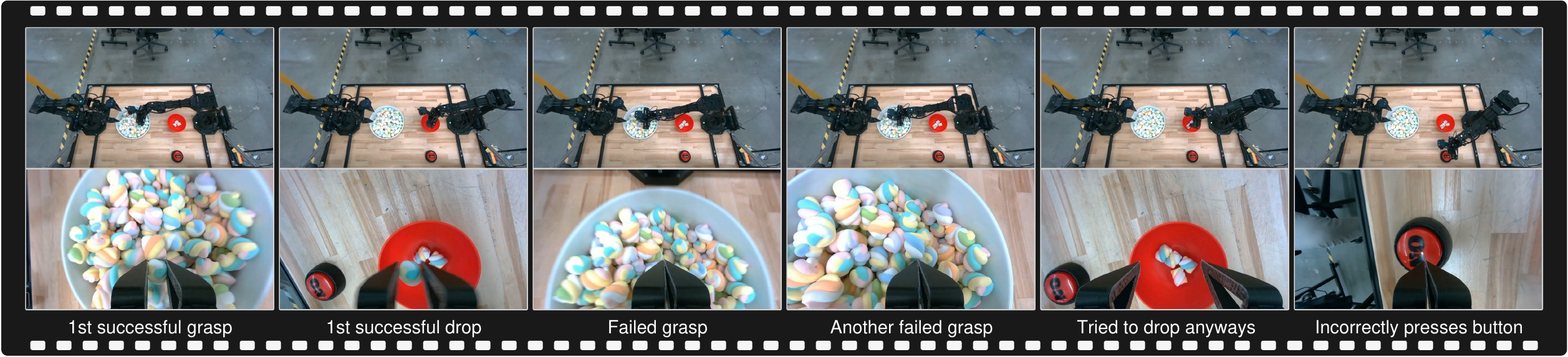} \\[6pt]
    \footnotesize\textbf{Current Obs (Failure)} \\[1pt]
    \includegraphics[height=\filmstripheight]{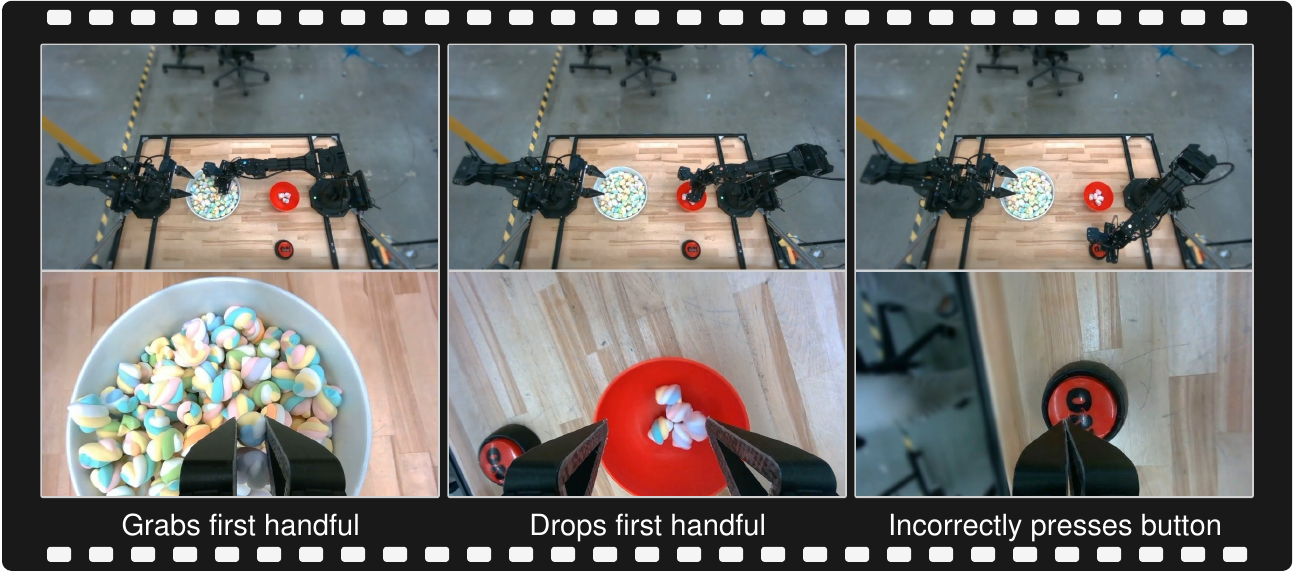}
    \end{tcolorbox}
    \caption{\textbf{Marshmallow evaluation behaviors.} Qualitative comparison of successful and failed executions on the marshmallow task.}
    \label{fig:filmstrip_marshmallow}
\end{figure}

\subsection{$\mathrm{Mug\ Replacement}$}
\begin{figure}[H]
    \centering
    \begin{tcolorbox}[
        width=0.95\textwidth,
        center,
        colback=figurebackground,
        boxrule=0pt,
        arc=10pt,
        left=6pt,
        right=6pt,
        top=4pt,
        bottom=4pt,
        boxsep=0pt,
        enhanced,
        colframe=figurebackground
    ]
    \centering
    \footnotesize\textbf{BPP (Success)} \\[1pt]
    \includegraphics[height=0.7\filmstripheight]{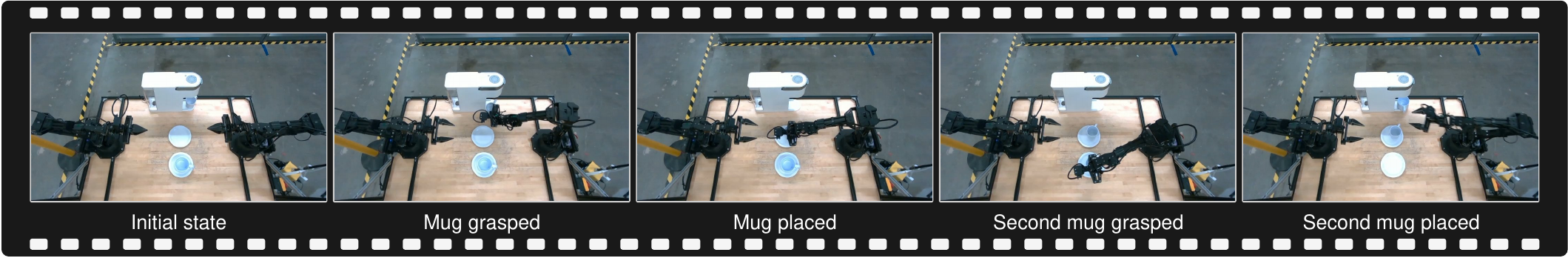} \\[6pt]
    \footnotesize\textbf{Na\"ive History (Failure)} \\[1pt]
    \includegraphics[height=0.7\filmstripheight]{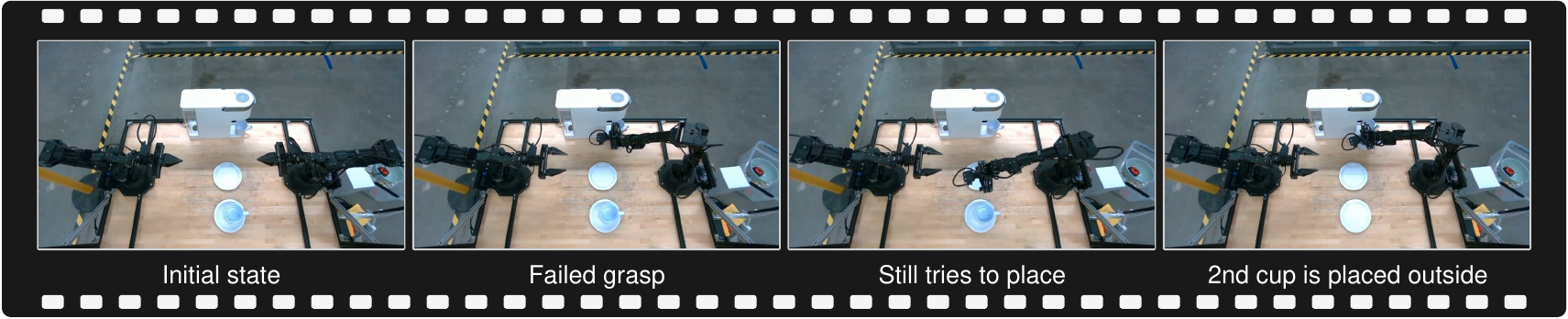}
    \end{tcolorbox}
    \caption{\textbf{$\mathrm{Mug\ Replacement}$ evaluation behaviors.} Comparison of BPP success versus Na\"ive History failure.}
    \label{fig:filmstrip_mug}
\end{figure}

\subsection{$\mathrm{Stacking\ Puzzle}$}
\begin{figure}[H]
    \centering
    \begin{tcolorbox}[
        width=0.95\textwidth,
        center,
        colback=figurebackground,
        boxrule=0pt,
        arc=10pt,
        left=6pt,
        right=6pt,
        top=4pt,
        bottom=4pt,
        boxsep=0pt,
        enhanced,
        colframe=figurebackground
    ]
    \centering
    \footnotesize\textbf{BPP (Success)} \\[1pt]
    \includegraphics[height=\filmstripheight]{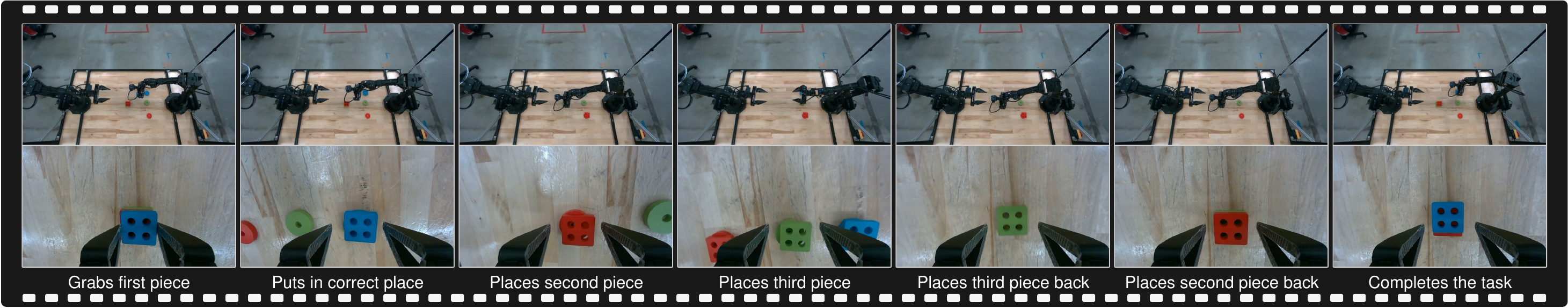} \\[6pt]
    \footnotesize\textbf{Na\"ive History (Failure)} \\[1pt]
    \includegraphics[height=\filmstripheight]{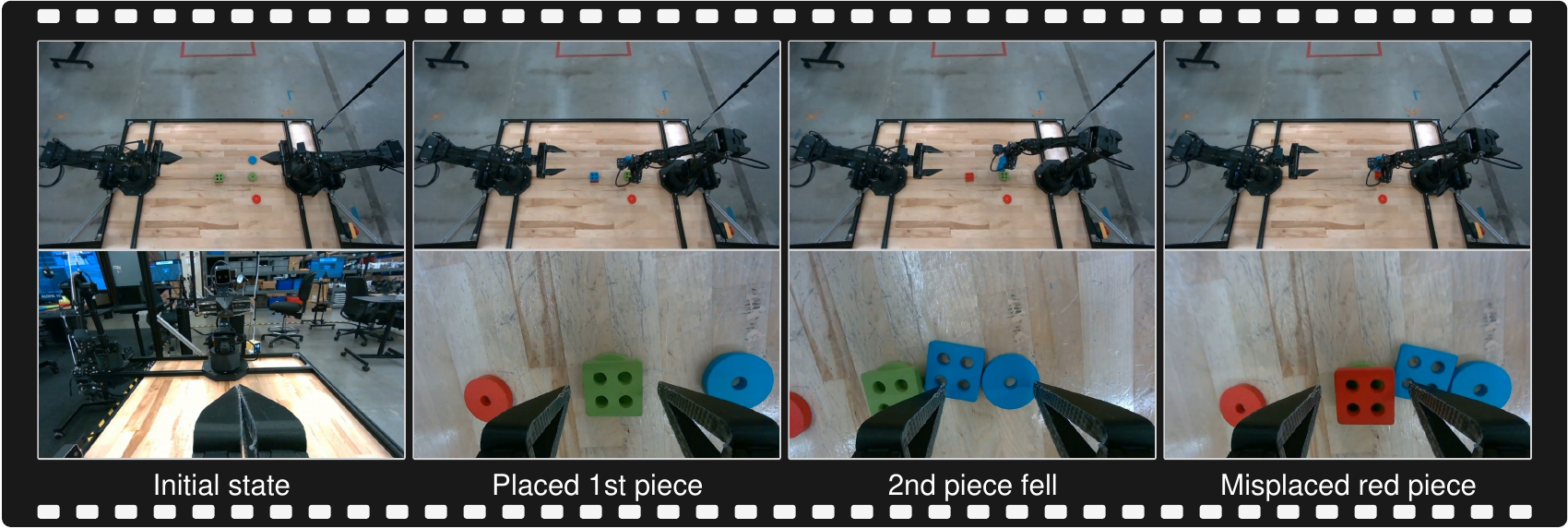}
    \end{tcolorbox}
    \caption{\textbf{$\mathrm{Stacking\ Puzzle}$ evaluation behaviors.} BPP method successfully solves the long-horizon sorting task.}
    \label{fig:filmstrip_stacking}
\end{figure}

\end{document}